\def\year{2022}\relax
\documentclass[letterpaper]{article} 
\usepackage{aaai22}  
\usepackage{times}  
\usepackage{helvet}  
\usepackage{courier}  
\usepackage[hyphens]{url}  
\usepackage{graphicx} 
\usepackage{natbib}  
\usepackage{caption} 
\DeclareCaptionStyle{ruled}{labelfont=normalfont,labelsep=colon,strut=off} 
\usepackage{algorithm}
\usepackage{algorithmic}
\usepackage{subfigure}
\usepackage{amsfonts}
\usepackage{amsmath}
\usepackage{amssymb}
\usepackage{newfloat}
\usepackage{listings}

\floatstyle{ruled}
\newfloat{listing}{tb}{lst}{}
\floatname{listing}{Listing}
\title{Semantically Contrastive Learning for Low-light Image Enhancement}
\author{
    Dong Liang\textsuperscript{\rm 1}, 
    Ling Li\textsuperscript{\rm 1},   
    Mingqiang Wei\textsuperscript{\rm 1}\thanks{Corresponding author.},  
    Shuo Yang\textsuperscript{\rm 2},
    Liyan Zhang\textsuperscript{\rm 1},  
    Wenhan Yang\textsuperscript{\rm 3},   
    Yun Du\textsuperscript{\rm 1},   
    Huiyu Zhou\textsuperscript{\rm 4}
}
\affiliations{
    \textsuperscript{\rm 1}Nanjing University of Aeronautics and Astronautics, MIIT Key Laboratory of Pattern Analysis and Machine Intelligence, Collaborative Innovation Center of Novel Software Technology and Industrialization\\
    \textsuperscript{\rm 2}vivo Mobile Communication~~~
    \textsuperscript{\rm 3}Nanyang Technological University~~~
    \textsuperscript{\rm 4}University of Leicester\\
    \{liangdong, liling, zhangliyan, muyun\}@nuaa.edu.cn, mingqiang.wei@gmail.com, shuo.yang@vivo.com, wenhan.yang@ntu.edu.sg,  hz143@leicester.ac.uk 

}

\usepackage{bibentry}

\begin{document}

\maketitle

\begin{abstract}
Low-light image enhancement (LLE) remains challenging due to the unfavorable prevailing low-contrast and weak-visibility problems of single RGB images. In this paper, we respond to the intriguing learning-related question -- if leveraging both accessible unpaired over/underexposed images and high-level semantic guidance, can improve the performance of cutting-edge LLE models? Here, we propose an effective semantically contrastive learning paradigm for LLE (namely SCL-LLE).
Beyond the existing LLE wisdom, it casts the image enhancement task as multi-task joint learning, where LLE is converted into three constraints of contrastive learning, semantic brightness consistency, and feature preservation for simultaneously ensuring the exposure, texture, and color consistency. SCL-LLE allows the LLE model to learn from \textit{unpaired} positives (normal-light)/negatives (over/underexposed), and enables it to interact with the scene semantics to regularize the image enhancement network, yet the interaction of high-level semantic knowledge and the low-level signal prior is seldom investigated in previous methods. Training on readily available open data, extensive experiments demonstrate that our method surpasses the state-of-the-arts LLE models over six independent cross-scenes datasets. Moreover, SCL-LLE's potential to benefit the downstream semantic segmentation under extremely dark conditions is discussed. 
\textit{Source Code: \url{https://github.com/LingLIx/SCL-LLE}}.
\end{abstract}

\begin{figure}[h]
\centering
\subfigure[Input]{
\includegraphics[width=2.72cm]{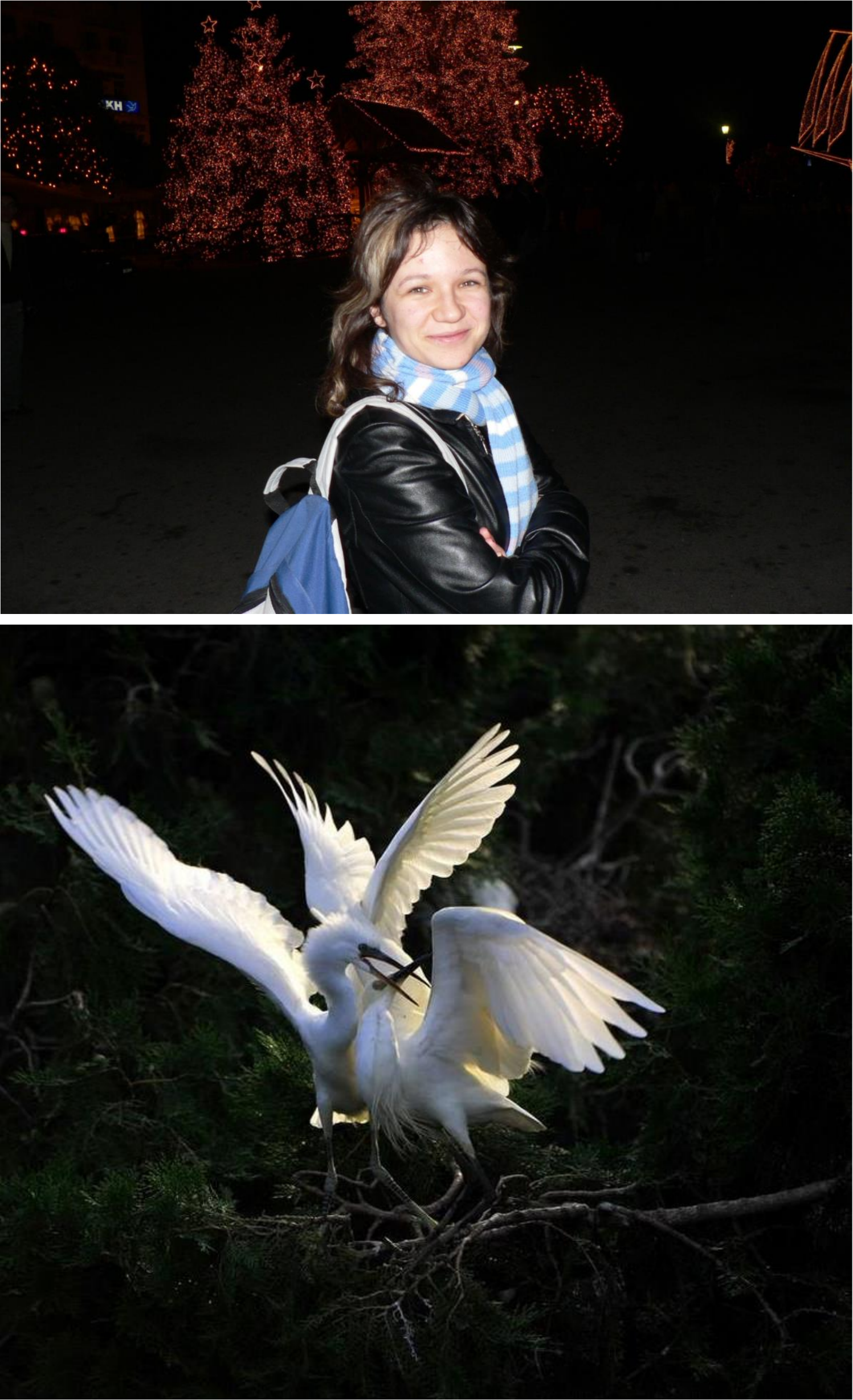}
}
\hspace{-0.3cm}
\subfigure[w/o Neg. samples]{
\includegraphics[width=2.72cm]{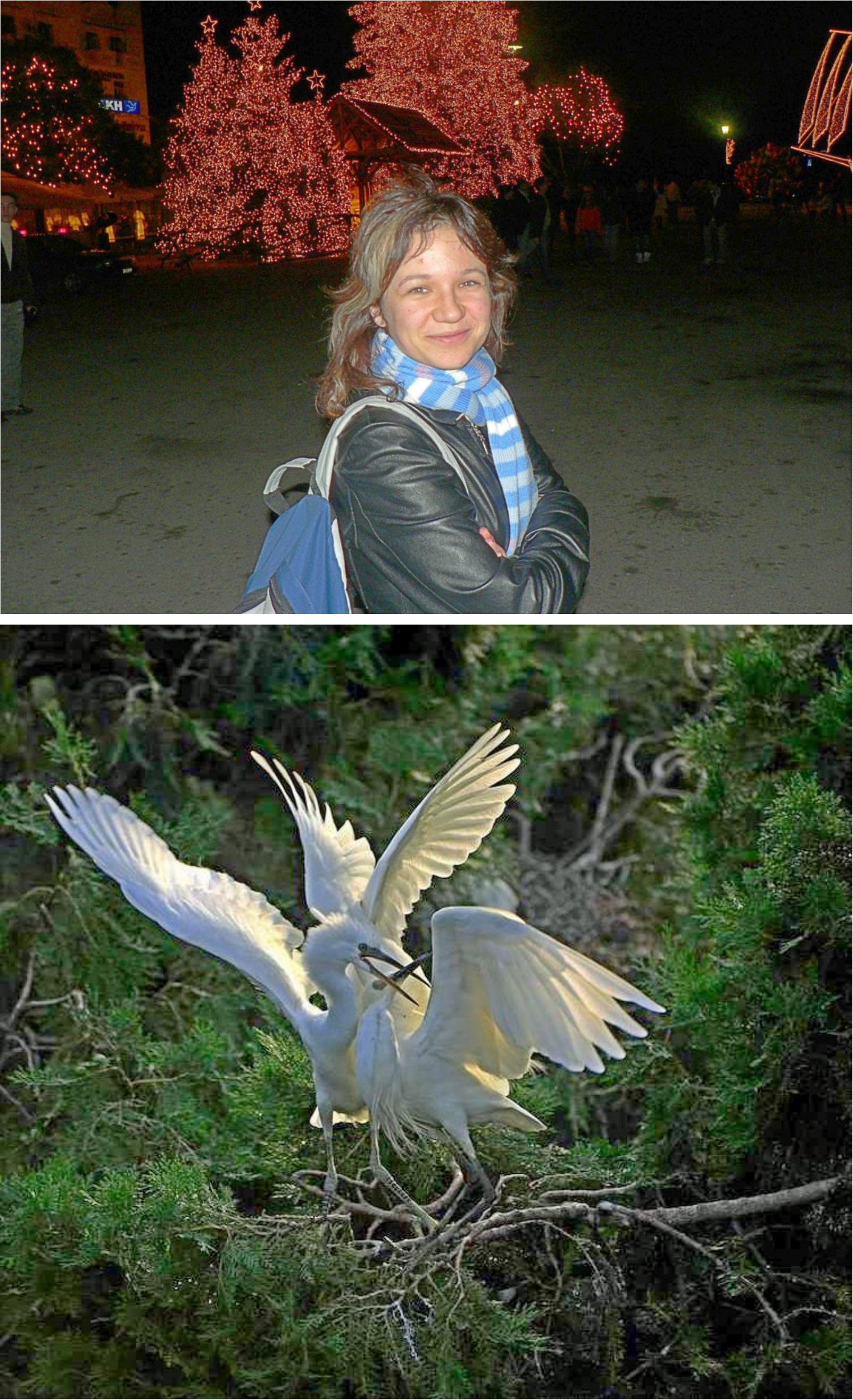}
}
\hspace{-0.3cm}
\subfigure[Ours]{
\includegraphics[width=2.72cm]{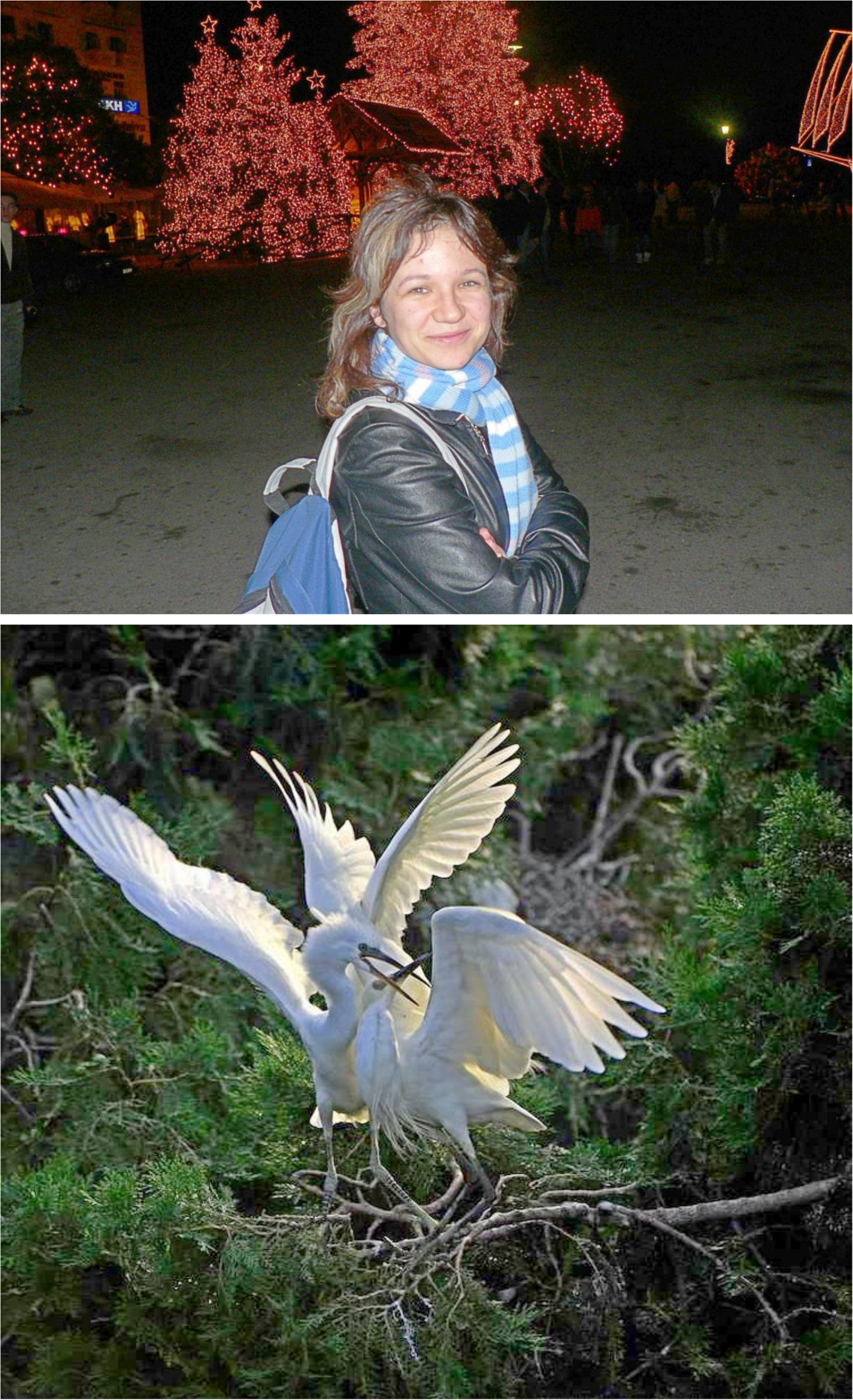}
}\\
\caption{Impact of training data. (b) no negative samples are involved in training.  
}
\label{img8}
\end{figure}

\section{Introduction}
\noindent Capturing images with underexposures remains a significant source of errors in camera imaging. 
Such degenerated images severely hinder some downstream tasks from operating smoothly, such as semantic segmentation or object detection in vision-based driving assistance systems.

Data-driven methods have been proposed to formulate the LLE problem by learning strict pixel correspondence of image {pairs} via strong supervisions, $e.g.$, \cite{lore2017llnet,Chen2018Retinex,zhang2019kindling,Xu_2020_CVPR,8692732}. Although having achieved remarkable successes in LLE, they only adopt the normal-illumination images as positive samples to guide the training while neglecting to exploit the over/underexposed images as negative samples. On the one hand, high-quality positive samples without brightness and color defects are challenging to acquire in practice. For example, Ignatov $et \ al.$ propose a scheme to acquire them from a DSLR camera to refine the imaging of a mobile phone camera~\cite{ignatov2017dslr}. It brings complicated registration and pixel-by-pixel calibration to image pairs. On the other hand, negative samples with error exposures are readily available. As shown in Fig.~\ref{img8}, our method yields visually better results, as the negative samples also provide supervisory information to generate more discriminative visual mapping.

Another drawback of current methods is the ignorance of the important semantic information. As shown in Fig.~\ref{DICM}, the enhancement strategies for the background and foreground should be different. Scene semantics can help to distinguish different areas for further enhancement. In addition, there is natural brightness consistency inside a semantic category. Employing such consistency can help to avoid local uneven exposure. Although methods~\cite{FanWY020,xie2020semantically} apply semantic information as guidance to improve image enhancement, they ignore the consistency between pixels of the same semantic category.

The motivation of this paper can be summarized as follows: 1) to release from the strict pixel correspondence of training image pairs and to leverage over/underexposed images as negative samples for building a more flexible and robust visual mapping method; 2) to effectively leverage the semantic information to distinguish enhancement areas and to keep brightness consistency inside the same semantic category. To this end, we develop a novel paradigm -- semantically contrastive enhancement learning for low-light image enhancement (SCL-LLE for short). SCL-LLE casts the image enhancement task as multi-task joint learning, where LLE is converted into three constraints of contrastive learning, semantic brightness consistency, and feature preservation for simultaneously ensuring the exposure, texture, and color consistency. 
From the respective of ease to use, we use readily accessible training data -- the Cityscapes~\cite{cordts2016cityscapes} dataset, to provide input images with semantic ground truths, and the Part1 of SICE dataset~\cite{Cai2018deep}, to provide unpaired negative/positive samples. 
The contributions of this paper are three folds: 
\begin{itemize}
    \item [$\bullet$] SCL-LLE removes pixel-correspond paired training data, and provides a more flexible way: 1) training with unpaired images in different real-world domains, 2) training with unpaired negative samples, allowing us to leverage readily available open data to build a more generalized and discriminative LLE network.
    \item [$\bullet$]
    Low and high-level vision tasks (i.e., LLE and semantic segmentation) promote each other. A semantic brightness consistency loss is introduced to ensure smooth and natural brightness recovery of the same semantic category.
    The enhanced images lead to better performance on the downstream semantic segmentation.
    \item [$\bullet$] SCL-LLE is compared with SOTAs via comprehensive experiments on six independent datasets, in terms of visual quality, no and full-referenced image quality assessment, and human subjective survey. All results consistently endorse the superiority of the proposed approach.
\end{itemize}

\begin{figure*}[!htb]
\centering
\includegraphics[width=15.5cm]{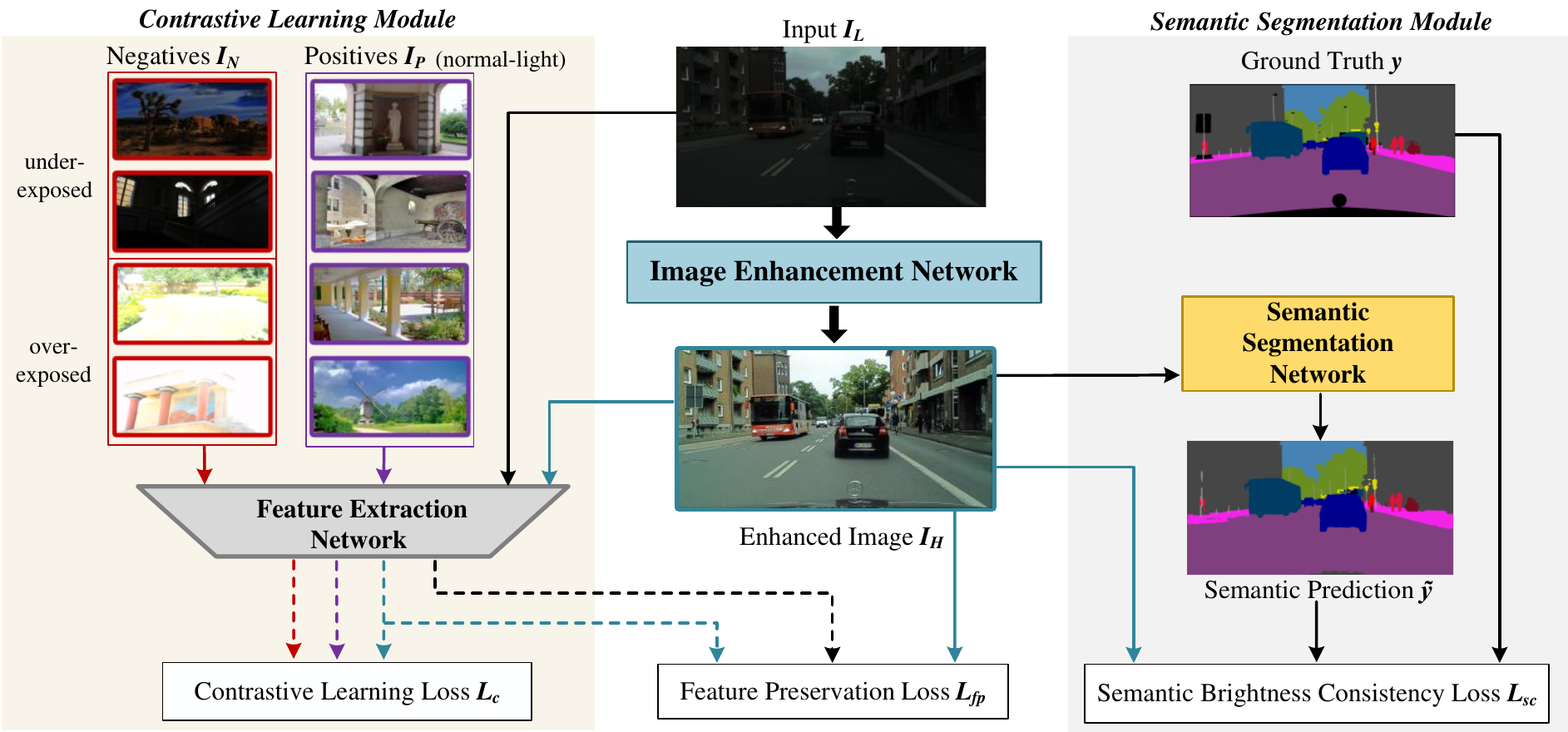}
\caption{Overall architecture of our proposed SCL-LLE. It includes a low-light image enhancement network, a contrastive learning module and a semantic segmentation module. 
}
\label{Net}
\end{figure*}
\section{Related Work}
\subsection{Low-light Image Enhancement}
Early efforts are commonly made towards the use of hand-crafted priors with empirical observations~\cite{Pizer1990ContrastlimitedAH,land1977retinex,xu2014novel,guo2016lime} to deal with the LLE problem. 
Histogram equalization ~\cite{Pizer1990ContrastlimitedAH}  used a cumulative distribution function to regularize the image's pixel values to distribute overall intensity levels evenly. Retinex model~\cite{land1977retinex} and its multi-scale version~\cite{Jobson1997A} decomposed the brightness into illumination and reflectance and then process them separately.
Guo~$et \ al.$ ~\cite{guo2016lime} introduced a structural prior to refine the initial illumination map and finally synthesized the enhanced image according to the Retinex theory. However, these handcrafted constraints/priors are not self-adaptive enough to recover image details and color, resulting in washing out details, local under/over-saturation, uneven exposure, or halo artifacts around objects.

Lore~$et \ al.$ ~\cite{lore2017llnet} proposed a variant of the stacked sparse denoising autoencoder to enhance the degraded images. RetinexNet~\cite{Chen2018Retinex} leveraged a deep architecture based on Retinex to enhance low-light images. Zhang~$et \ a l.$~\cite{zhang2019kindling} developed three subnetworks for layer decomposition, reflectance restoration, and illumination adjustment based on Retinex. RUAS~\cite{liu2021retinex} constructed the overall LLE network architecture by unfolding its optimization process. 
The above methods are trained based on image pairs with strict pixel correspondence. 

Recently, \cite{jiang2021enlightengan} reported an unsupervised method using normal-light images which do not have low-light images as correspondences.  Zero-DCE~\cite{Guo_2020_CVPR} reformulated the LLE task as an image-specific curve estimation problem with a fixed default brightness value without any reference image. Different from~ \cite{jiang2021enlightengan} and \cite{Guo_2020_CVPR}, our method learns the appropriate brightness from the unpaired negative samples. 

In another recent work, Fan~$et \ al.$ ~\cite{FanWY020} used semantic information to guide the reconstruction of the reflection of Retinex to eliminate noise. In contrast, we directly introduce semantic information to the brightness reconstruction and pay more attention to the dependency among the semantic elements via the interaction of high-level semantic knowledge and low-level signal priors. 
\subsection{Contrastive Learning}
Contrastive learning~\cite{he2020momentum,chen2020simple,sermanet2018time,tian2019contrastive,henaff2020data} is from the self-supervised learning paradigm, which is characterized by using pretext tasks to mine its own supervisory information from original data for downstream tasks. 
For a given input, contrastive learning aims to pull it together with the positives and push it apart from negatives in a representation space. 
Previous works have applied contrastive learning to high-level vision tasks because these tasks are inherently suited for modeling the contrast between positive and negative samples. A recent work~\cite{wu2021contrastive} has demonstrated that it can improve image dehaze. 

In this paper, we present contrastive learning to remove the dependency of training data and allows us to use both positive and negative samples, helping to collect training samples from available open data. In addition, most of the existing contrastive learning methods rely heavily on a large number of negative samples and thus require either large batches or memory banks \cite{li2021triplet}. In our approach, we employ only a couple of negative samples for one positive sample and introduce a random mapping strategy to avoid the risk of overfitting.
\section{Methodology}
\subsection{Problem Formulation and Overall Architecture}
Fundamentally, low-light image enhancement can be regarded as seeking a mapping function ${F}$, such that $I_H = {F}(I_L)$ is the desired image, which is enhanced from the input image $I_L$. In our design, we introduce two different priors: one is the contrastive samples including the negatives $I_N$, i.e., the under/overexposed images, and the positives $I_P$, i.e., the normal-light images; the other is the semantic-related priors including the semantic segmentation ground truth $y$ and the semantic prediction $\tilde{y}$. Therefore, we formulate a new mapping function as follows:
\begin{equation}
    I_H = {F}(I_L, I_N, I_P, y, \tilde{y})
\end{equation}
At the top level, we design a novel semantically contrastive learning framework for LLE (call SCL-LLE) as shown in Fig.~\ref{Net} to obtain a better mapping function $F$ than the existing methods do. 

SCL-LLE consists of an image enhancement network, a semantic segmentation network, and a feature extraction network. Specifically, given an $I_L$ input, the image enhancement network is first applied, and the enhanced result is then fed into the following semantic segmentation network. For the task modules, we leverage three mainstream networks: The low-light image enhancement network is a U-Net like backbone~\cite{Guo_2020_CVPR}, which remaps every pixel by generating the pixel correction curve; the semantic segmentation network that we use here is the popular DeepLabv3+~\cite{chen2018encoder}; the feature extraction network is the VGG-16~\cite{simonyan2014very}. We use three losses to correspond to the three aspects -- Contrastive learning, semantic brightness consistency, and feature preservation. 

\subsection{Contrastive Brightness Restoration}
We use the normal-light and over/underexposed images as the positive samples and negative samples to restore the brightness of the low-light images. 
Note that, to ensure the flexibility of the method, the positive and negative samples can be select in different scenes with each other and with the input image, i.e., they are unpaired with the input image and also unpaired with each other. 

The goal is to learn a representation to pull together “positive” pairs in the latent feature space and push apart the representation between “negative” pairs. We need to consider two aspects: to construct the “positive” pairs and “negative” pairs, and to find the latent feature space of these pairs for contrasting. In our method, the positive pair is generated by a normal-light image $I_{P}$ and an enhanced image $I_H$ by the low-light image enhancement network. Similarly, the negative pair is generated by  an over/underexposure image $I_{N}$ and an enhanced image $I_H$.
For the latent feature space, besides using the appearance of the image, we choose a pre-trained VGG-16 to extract the feature map $f\in \mathbb{R}^{C\times H\times W}$, where $G^{l}_{ij}$ is the inner product between the feature maps $i$ and $j$ in the layer $l$:
\begin{equation}
    G_{i j}^{l}=\sum_{k} f_{i k}^{l} f_{j k}^{l}
\end{equation}
where $k$ represents the vector length. 
We then get a set of Gram matrices $\left\{G^{1}, G^{2}, \ldots, G^{L}\right\}$ from layers $1, . . . , L$ in the feature extraction network.
The Gram matrix $G$ is a quantitative description of latent image features. Similar to the triplet loss~\cite{schroff2015facenet,2017In}, our goal is:
\begin{equation}
  d(G(I_{H}), G(I_{P}))\ll d(G(I_{H}), G(I_{N}))
\end{equation}
\begin{equation}
  d(E(I_{H}), E(I_{P}))\ll d(E(I_{H}), E(I_{N}))
\end{equation}
where $E$ represents the expectation. We wish that the distance $d$ between features $I_{H}$ and $I_{P}$ is smaller than the distance between features $I_{H}$ and $I_{N}$. The contrastive learning loss $L_{c}$ can be expressed as:
\begin{equation}
\begin{aligned}
  L_{c}&=max\left \{ d(G(I_{H}), G(I_{P}))-d(G(I_{H}), G(I_{N})) + \alpha , 0 \right \} \\
  &+max\left \{ d(E(I_{H}), E(I_{P}))-d(E(I_{H}), E(I_{N})) + \beta , 0 \right \}
\end{aligned}
\end{equation}
where $\alpha$ and $\beta$ are hyperparameters (i.e., the margin in the triplet loss), we set them to 0.3 and 0.04 respectively in our experiments. In our implementation, the triplet loss is used to formulate the above loss function, which is a particular case of the contrastive loss when the number of the positive and negative samples is one. The details of the loss derivation can be found in supplementary material.

In the latest theoretical work \cite{li2021triplet}, the authors argued that negative pairs are
necessary, but one is sufficient for a triplet loss. They also observe that contrastive learning for visual representation can gain significantly from randomness. Inspired by this work, to collect positive and negative samples, we leverage SICE~\cite{Cai2018deep} dataset, which contains low-contrast images with different exposure levels and their corresponding high-quality images. The SICE dataset includes indoor/outdoor 589 scenes (Part1 with 360 scenes and Part2 with 229 scenes) with a total number of 4413 multi-exposure images. All 360 standard images in all the scenes of Part1 are used as positive samples. We choose one underexposed and one overexposed sample as negatives for each scene, a total of 720 images from Part1 of SICE. To improve the robustness of the model, during training, positive and negative samples are randomly selected in each iteration.  
\subsection{Semantic Brightness Consistency Constraint}
To better maintain the details of the image and make full use of its semantic information, we propose a semantic consistency loss to restrict brightness consistency and smoothness. This constraint can ensure that the same semantic parts in the enhanced image are consistent, which proves to be critical in avoiding local over/underexposures, as the experiments reveal later on.

In a real scene, the elements belonging to the same semantic category have a clustered or adjacent location and should be similar with consistent brightness ($e.g.$, the sky and the road). The existing enhancement networks cannot make the parts of inconsistent brightness to be smooth. Based on this observation, we define an average value $B$ of the brightness level of the overall pixels in each semantic category as follows:
\begin{equation}
  B_{s}=\frac{1}{n}\sum_{i\in {\theta_{s}}}^{}(B_{I_{H}}^{i})
\end{equation}  
where $s$ represents the $s$-th category, and we can attain multiple averages representing individual categories separately $\left\{B_{1}, B_{2}, \ldots \right\}$. $n$ represents the number of the semantic pixels in this category with the ground truth $y$. We denote $\theta_{s}$ as the pixel index collection belonging to category $s$, $B_{I_{H}}^{i}$ as the brightness level in the enhanced image $I_{H}$ at the category $s$. The semantic brightness consistency loss $L_{sc}$ is defined as:
\begin{equation}
  L_{sc}=\sum_{s=1}^{S}\sum_{i\in {\theta_{s}}}^{}(B_{I_{H}}^{i}-B_{s})^{2}-\sum_{s=1}^{S}(p_{s}*\log q_{s})
\end{equation}
where $S$ is the number of the categories of the semantic prediction $\tilde{y}$, $p_{s}$ represents the ground truth value, and $q_{s}$ represents the predicted value in the semantic prediction $\tilde{y}$.

To collect input images with semantic ground truths, we use the Cityscapes \cite{cordts2016cityscapes} dataset, which is a street scene dataset and contains 5000 images with pixel-level fine/coarse annotations. There are 2975 images for training, 500 for validation, and 1525 for testing. We train our network on the training images of the Cityscapes dataset for two reasons. First, it contains the fine semantic annotation information needed in our experiments. Second, Cityscapes captured street scenes with a low-end sensor (1/3 in CMOS 2MP), having the prevailing low-quality imaging samples.
We collect all the samples with fine annotations from the low dynamic range (LDR) training set, including various lighting scenes. We further adjust the brightness of these images to augment more severe no normal-light imaging conditions. 
\subsection{Feature Preservation}
Many low-level visual tasks~\cite{ledig2017photo,Kupyn_2018_CVPR,johnson2016perceptual} use the perceptual loss to make desired images and their ground truth perceptually consistent. To be distinct from the typical usage of the perceptual loss, we reformulate it as a feature retention loss to preserve the image features before and after enhancement. The feature retention loss $L_{fr}$ is defined as:
\begin{equation}
L_{fr}=\frac{1}{C_{l}W_{l}H_{l}}(f^{l}(I_{L})-f^{l}(I_{H}))^{2}
\end{equation}
where $f^{l}(I_{L})$ denotes the feature map $f\in \mathbb{R}^{C\times H\times W}$ of the input image $I_{L}$ in the layer $l$, and $f^{l}(L_{H})$ is the feature map of the enhanced image $I_{H}$ in the layer $l$.

Since the color naturalness is one of the significant concerns of LLE, we add a color constancy term $L_{cc}$ incorporating with the feature retention term, following the way reported in \cite{Guo_2020_CVPR}. It is based on the gray-world color constancy hypothesis \cite{buchsbaum1980spatial} that the pixel averages of the three channels tend to be of the same value. $L_{cc}$ constrains the ratio of three channels to prevent potential color deviations in the enhanced image. In addition, to avoid aggressive and sharp changes between neighboring pixels, an illumination smoothness penalty term is also embedded in $L_{cc}$. The formulation of $L_{cc}$ can be expressed as:
\begin{equation}
\begin{aligned}
    L_{cc} &=\sum_{\forall (p,q)\in \xi }(J^p-J^q)^2 \\
            &+ \lambda \frac{1}{M}\sum_{m=1}^{M}\sum_{p\in \xi }(\left |{\triangledown _x{A}_{m}^{p}} \right|+\left |{\triangledown _y{A}_{m}^{p}} \right|), \xi=\left \{ R,G,B \right \}
\end{aligned}
\end{equation}
where $J^p$ denotes the average intensity value of $p$ channel in the enhanced image, $(p,q)$ represents a pair of channels, $M$ is the number of the iterations, and $\triangledown _x$ and $\triangledown _y$ denote the horizontal and vertical gradient operations, respectively. The curve parameter map $A$ \cite{Guo_2020_CVPR} is the output at each iteration. We set $\lambda$ to 200 in our experiments for the best outcome.
The feature preservation loss $L_{fp}$ is the sum of $L_{fr}$ and $L_{cc}$.
\renewcommand{\thefootnote}{\fnsymbol{footnote}}
\begin{figure*}[!htb]
\centering
\subfigure[Input]{
\includegraphics[width=3.9cm]{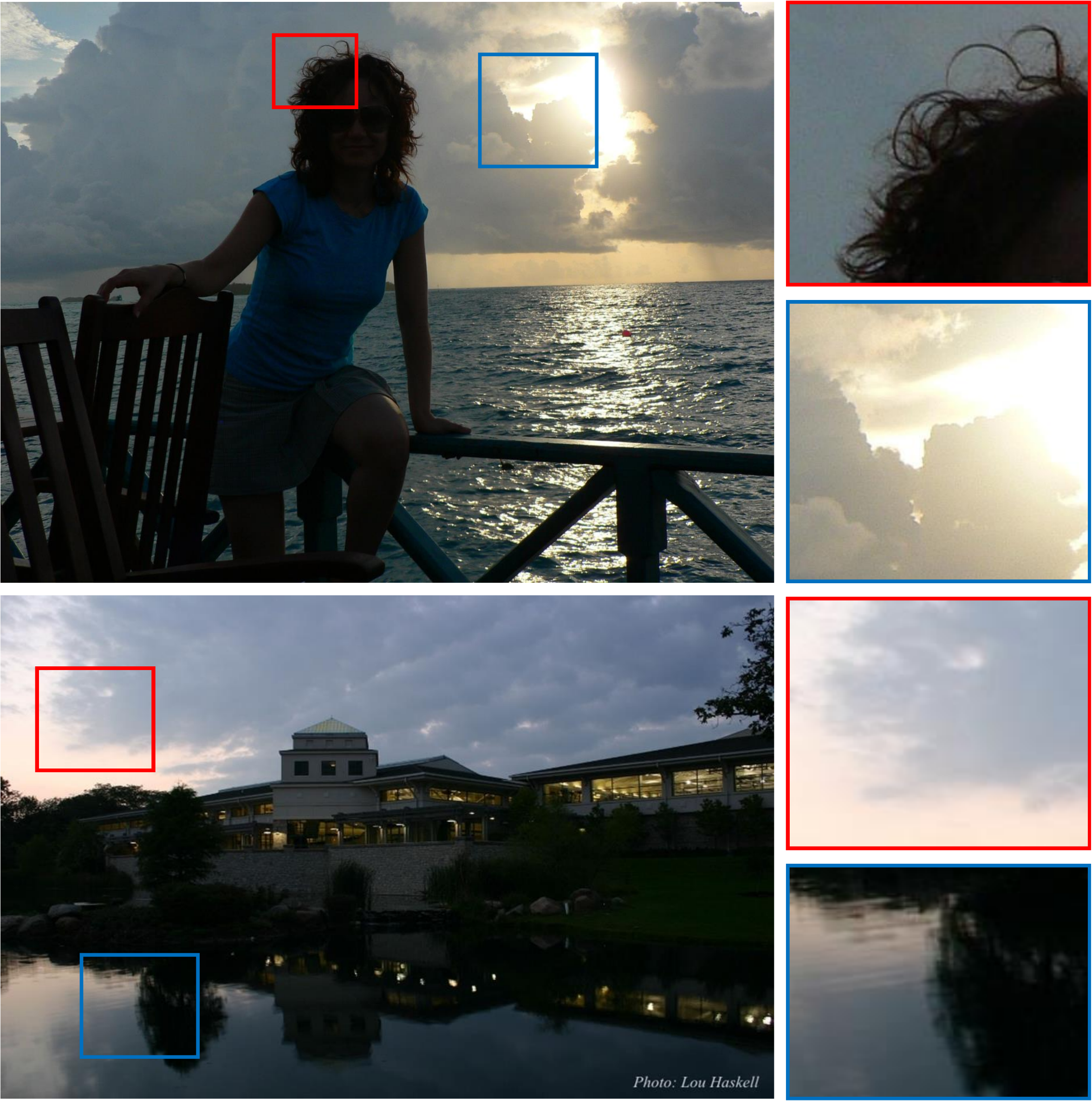}
}\hspace{-1mm}
\subfigure[LIME]{
\includegraphics[width=3.9cm]{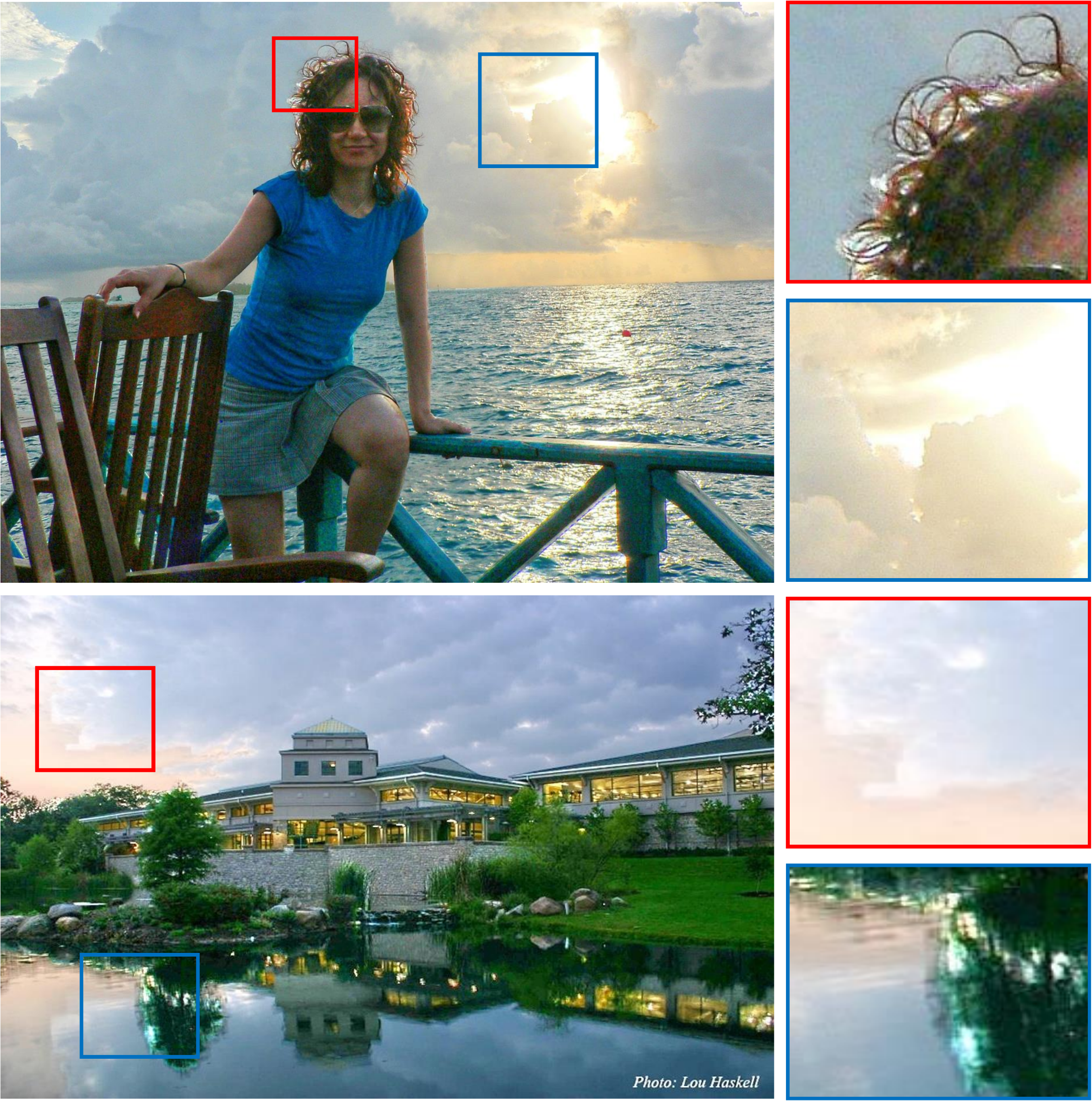}
}\hspace{-1mm}
\subfigure[RetinexNet]{
\includegraphics[width=3.9cm]{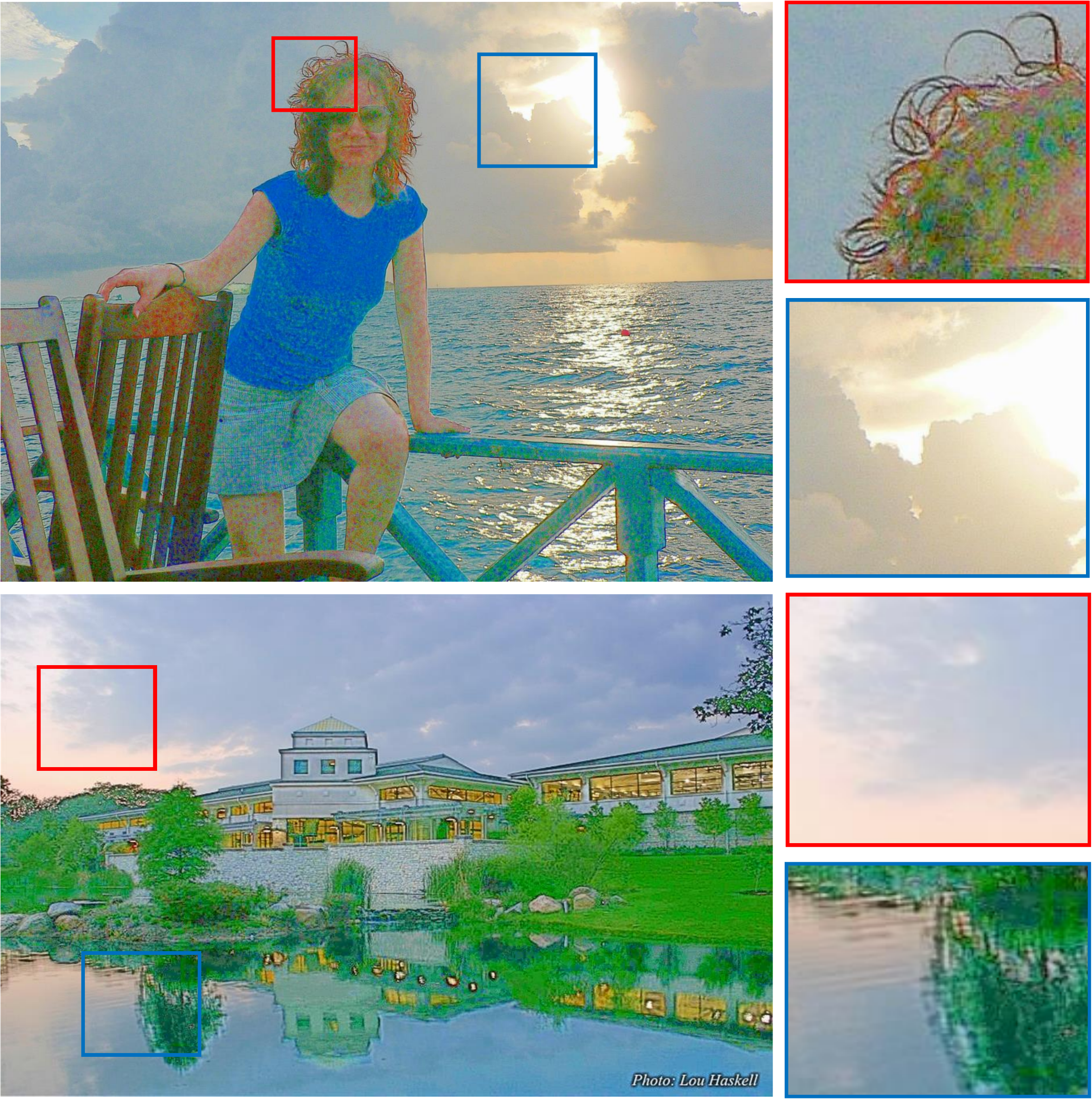}
}\hspace{-1mm}
\subfigure[ISSR]{
\includegraphics[width=3.9cm]{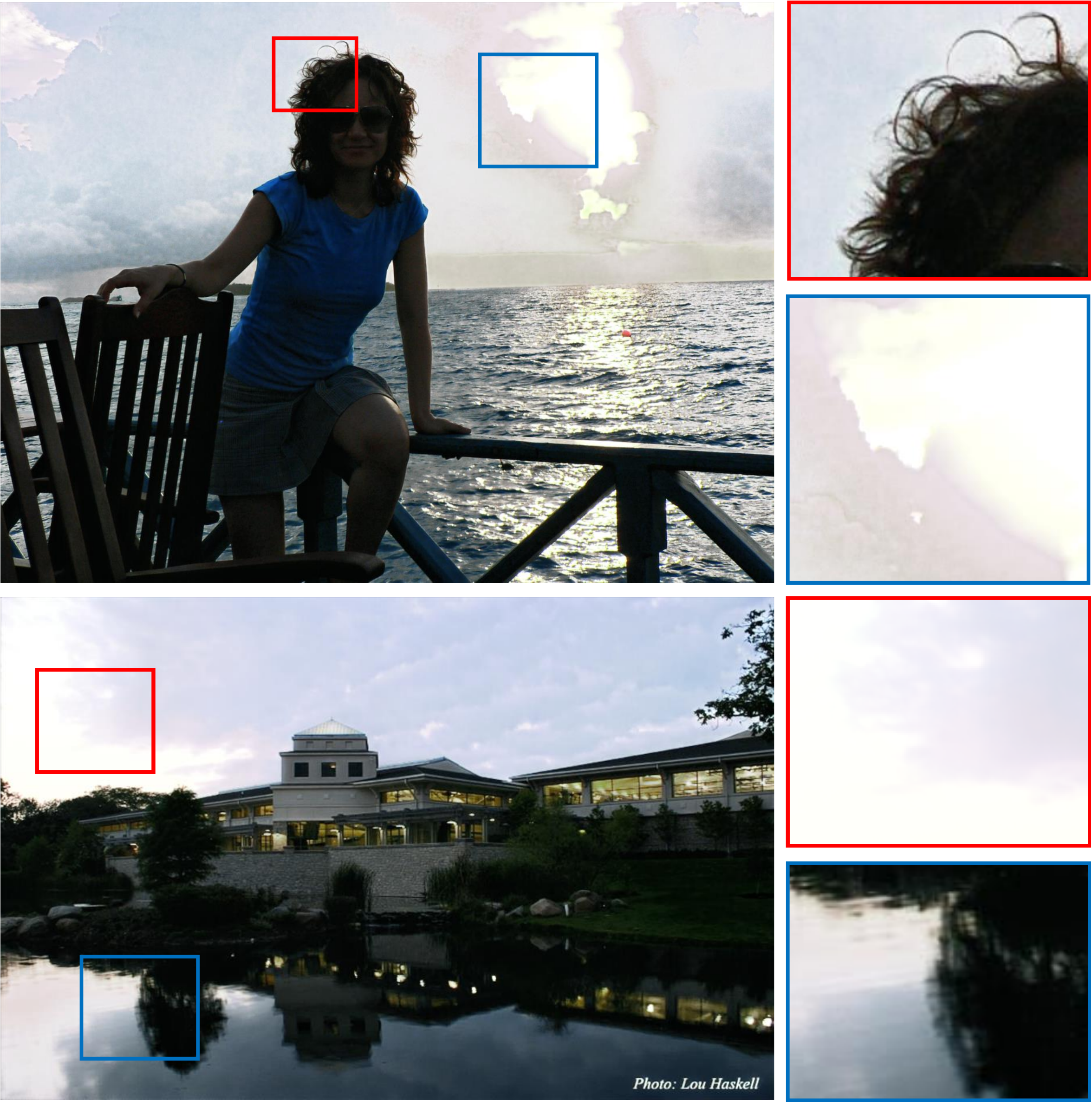}
}
\quad
\subfigure[Zero-DCE]{
\includegraphics[width=3.9cm]{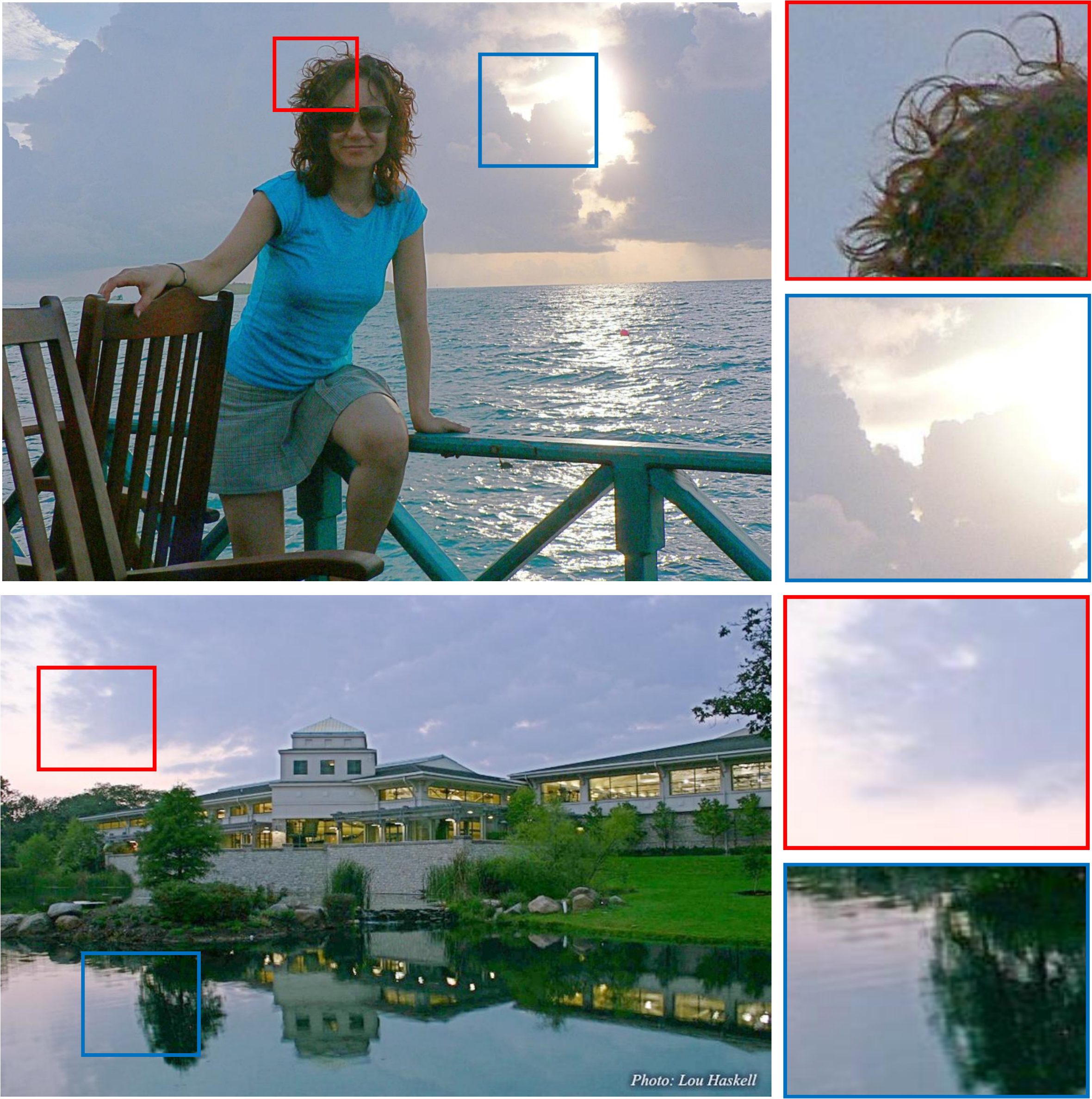}
}\hspace{-1mm}
\subfigure[EnlightenGAN]{
\includegraphics[width=3.9cm]{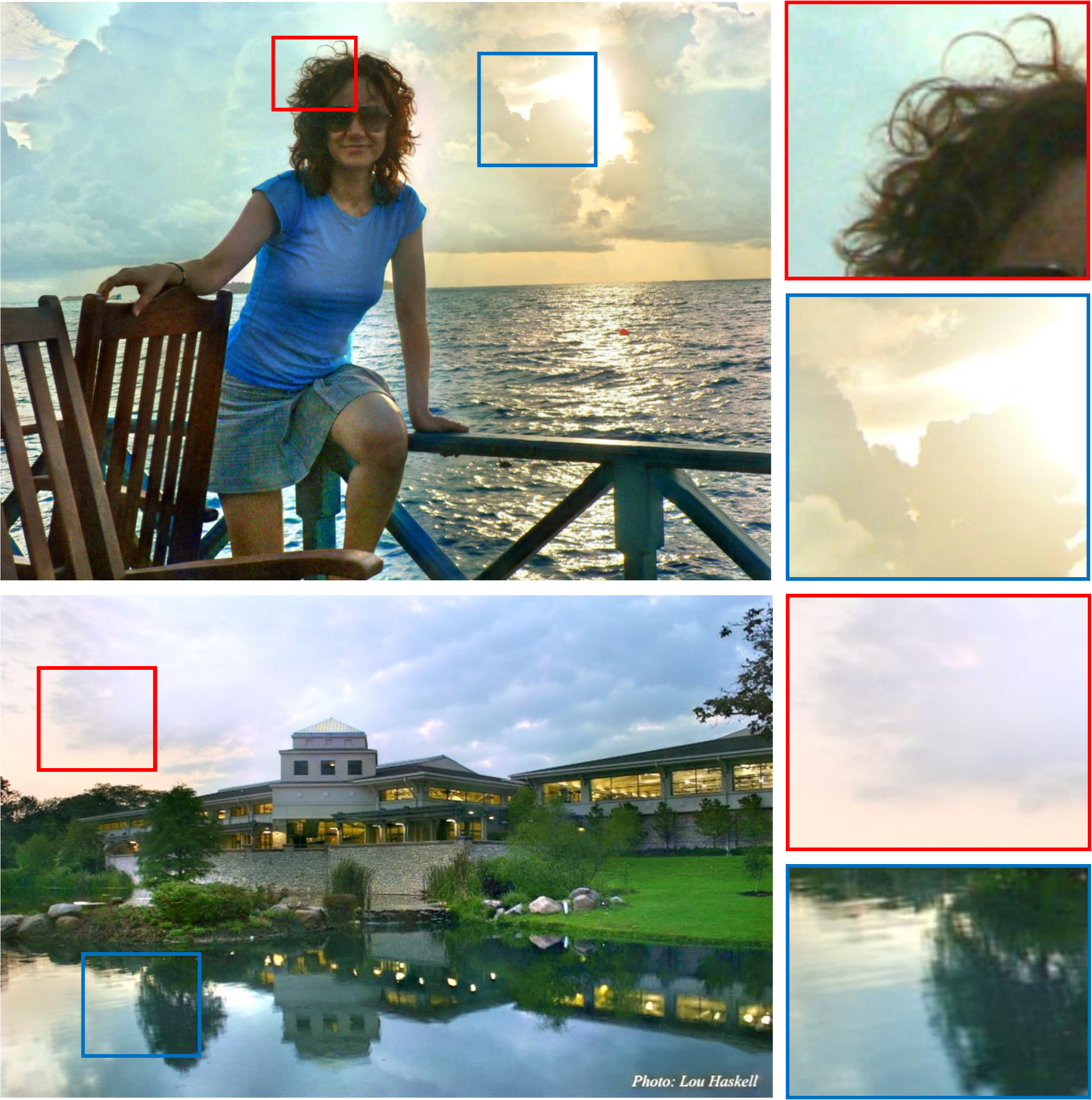}
}\hspace{-1mm}
\subfigure[RUAS]{
\includegraphics[width=3.9cm]{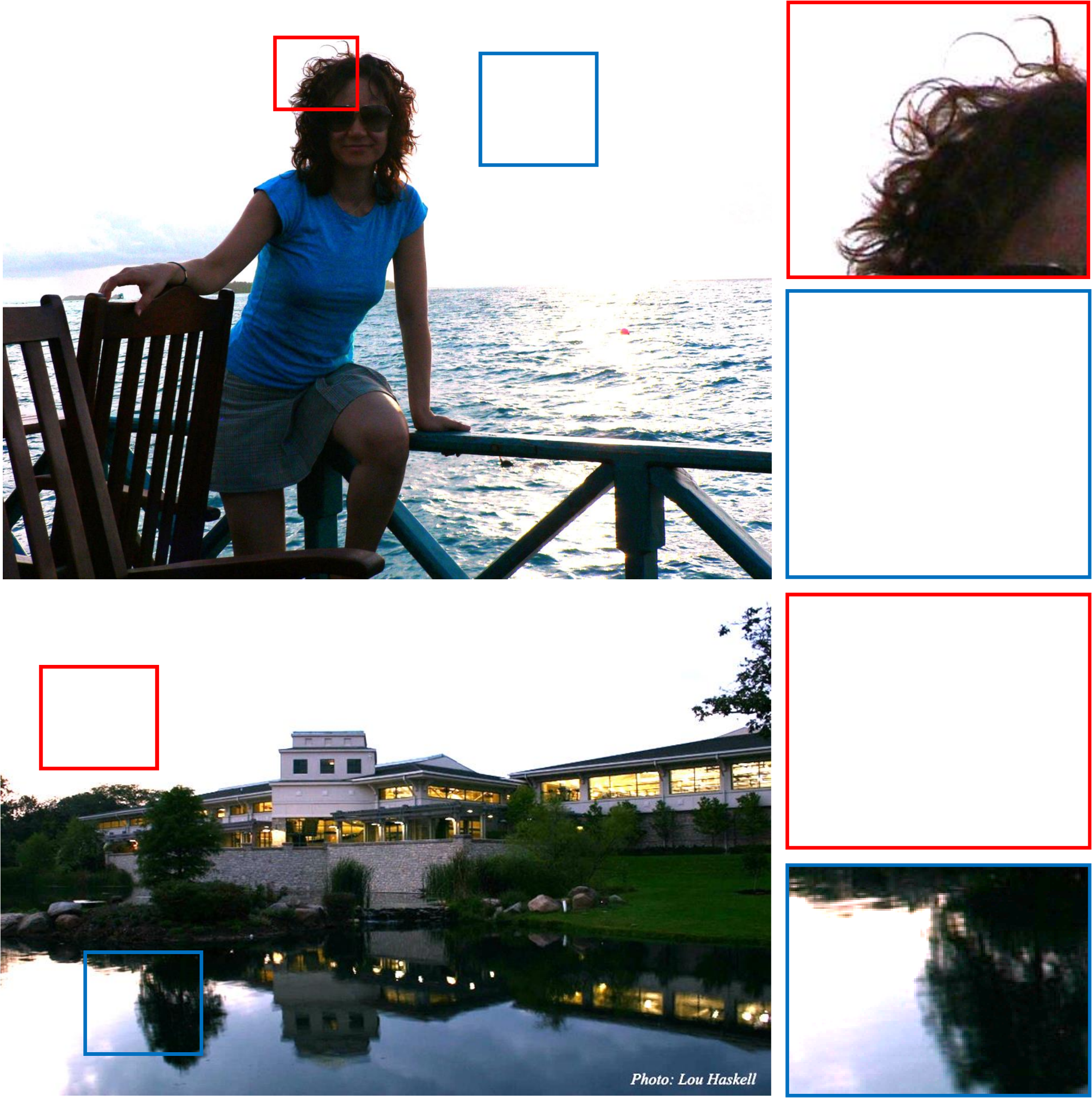}
}\hspace{-1mm}
\subfigure[Ours]{
\includegraphics[width=3.9cm]{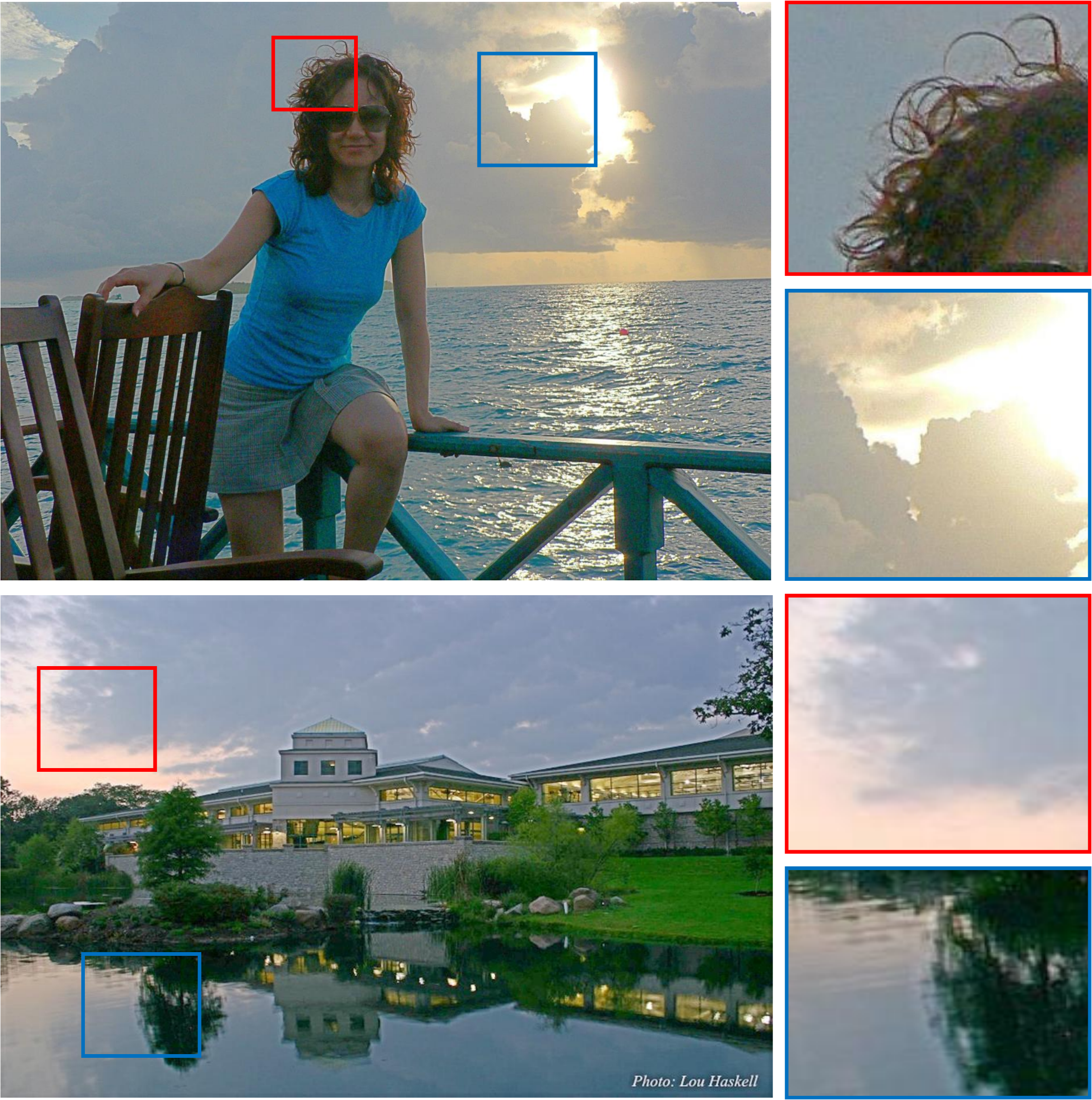}
}\\
\caption{Comparison of SCL-LLE and the state-of-the-art methods over VV and DICM datasets with zoom-in regions.
Our method enables the enhanced images to look more realistic, and recovers better details in both foreground and background.  
}
\label{DICM}
\end{figure*}
\subsection{Efficient Training Details} 
In our implementation, the semantic segmentation network is pre-trained on the Cityscapes dataset, and the feature extraction network is pre-trained on ImageNet~\cite{ILSVRC15}. We train SCL-LLE end-to-end while fixing the weights of the semantic segmentation network and the feature extraction network. The back-propagated operation updates only the weights in the image enhancement network. Hence, most network computation is done in the image enhancement network, which efficiently learns $I_H$ from $(I_L, I_N, I_P, y, \tilde{y})$ to recover the enhanced image with a wide variety of scenes and semantics. 
We resize the training images to the size of 384×384. As for the numerical parameters, we set the maximum epoch as 50 and the batch size as 2. Our network is implemented with PyTroch on an NVIDIA 1080Ti GPU. The model is optimized using the Adam optimizer with a fixed learning rate of $1e^{-4}$.
\begin{table*}[!htb]
\centering
\caption{NIQE $\downarrow$, UNIQUE (UN.) $\uparrow$ and User Study (U.S.) $\downarrow$ scores on DICM, LIME, MEF, VV, and NPE datasets. 
}
\resizebox{2.1\columnwidth}{!}
{
\begin{tabular}{c|ccc|ccc|ccc|ccc|ccc|ccc}
\hline
 &\multicolumn{3}{c|}{DICM} &\multicolumn{3}{c|}{LIME} &\multicolumn{3}{c|}{MEF} &\multicolumn{3}{c|}{VV} &\multicolumn{3}{c|}{NPE} &\multicolumn{3}{c}{Average} \\ 
\cline{2-19}
Methods&NIQE $\downarrow$&UN. $\uparrow$&U.S. $\downarrow$&NIQE $\downarrow$&UN. $\uparrow$&U.S. $\downarrow$&NIQE $\downarrow$&UN. $\uparrow$&U.S. $\downarrow$&NIQE $\downarrow$&UN. $\uparrow$&U.S. $\downarrow$&NIQE $\downarrow$&UN. $\uparrow$&U.S. $\downarrow$&NIQE $\downarrow$&UN. $\uparrow$&U.S. $\downarrow$\\ \hline
Input &4.26 &0.72 &3.33 &4.36 &0.70 &4.30 &4.26 &0.72 &4.41 &3.52 &\textbf{0.74} &3.38 &4.32 &\textbf{1.17} &3.92 &4.13 &0.75 &3.67 \\ 
(TIP'17) LIME &3.75 &0.78 &3.44 &3.85 &0.53 &{2.10}& 3.65 &0.65 &3.82 &\textbf{2.54} &0.44 &2.75 &4.44 &0.93 &3.75&3.55 &0.69 &3.40 \\ 
(BMVC'18) R.-Net &4.47 &0.75 &3.59 &4.60 &0.52 &4.00 &4.41 &0.97 &4.06 &{2.70} &0.36 &2.88 &4.60 &0.81 &4.13 &4.13 &0.69 &3.75 \\
(ACMMM'20) ISSR &4.14 &0.59 &3.13 &4.17 &\textbf{0.83} &3.40 & 4.22 &0.87 &4.47 & 3.57 &0.62 &3.00 & 4.02 &0.99 &3.96 &4.03 &0.68 &3.49\\
(CVPR'20) Z.-DCE &3.56 &0.82 &2.77 &{3.77} &0.73 &{2.10}& {3.28} &1.22 &{3.18} &3.21 &0.48 &2.50 &{3.93} &1.07 &{2.50}&3.50 &0.81 &{2.70} \\
(TIP'21) E.GAN &{3.55} &0.63 &2.81 &\textbf{3.70} &0.49 &\textbf{2.00} & \textbf{3.16} &1.03 &3.29 & 3.25 &0.58 &{2.12}& 3.95 &1.07 &2.85 &{3.47} &0.69 &2.72\\
(CVPR'21) RUAS &5.21 &-0.17 &3.44 &4.26 &0.34 &2.30& 3.83 &0.73 &4.11 &4.29 &-0.04 &3.75 &5.53 &0.13 &4.17 &4.78 &0.04 &3.60 \\    \hline     
Ours &\textbf{3.51} &\textbf{0.87} &\textbf{2.73}& 3.78 &0.76 &2.20& 3.31 &\textbf{1.25} &\textbf{2.47} &3.16 &0.49 &\textbf{1.63} &\textbf{3.88} &1.08 &\textbf{2.08}&\textbf{3.46} &\textbf{0.85} &\textbf{2.46}\\ 
\hline
\end{tabular}}
\label{tab1}
\end{table*}


\section{Experiments}
\subsection{Cross-dataset Peer Comparison}
For testing images, we use six publicly available low-light image datasets from other reported works, i.e., DICM~\cite{lee2012contrast}, MEF~\cite{ma2015perceptual}, LIME~\cite{guo2016lime}, NPE~\cite{wang2013naturalness}, VV\footnote{https://sites.google.com/site/vonikakis/datasets} and the Part2 of SICE~\cite{Cai2018deep}). 
Note that all the images in the six datasets are independent cross-scenes images without any overlapped scene of the input image and the positive/negative samples.

We compare the proposed method with six representative heterogeneous state-of-the-art methods, including a latest
conventional method LIME~\cite{guo2016lime};
a GAN-based method EnlightenGAN~\cite{jiang2021enlightengan}; three Retinex-based methods  RetinexNet~\cite{Chen2018Retinex},
RUAS~\cite{liu2021retinex}, and ISSR~\cite{FanWY020}, where ISSR also leverages semantic knowledge; and Zero-DCE~\cite{Guo_2020_CVPR} which leverages the same backbone enhancement network with ours. The results of the above methods are reproduced by the publicly available models provided with the recommended test settings.
\subsubsection{Visual quality comparison.}
We first examine whether the proposed methods can achieve visually pleasing results in brightness, color, contrast, and naturalness. We observe from Fig.~\ref{DICM} that all the SOTAs sacrifice over/under/uneven exposure in global or local areas. Specifically, LIME leads to color artifacts in strong local edges ($e.g.$, hair and sky, and inverted reflection in the water); RetinexNet and EnlightenGAN cause global color distortions with details
missing; ISSR and RUAS generate severe global and local over/underexposure. In contrast, our method recovers more details and better contrast in both foreground and background, thus enabling the enhanced images to look more realistic with vivid and natural color mapping. 
\subsubsection{No-referenced image quality assessment.}
We adopt Natural Image Quality Evaluator (NIQE)~\cite{Mittal2013MakingA}, a well-known no-reference image quality assessment for evaluating image restoration without ground-truth and providing quantitative comparisons. Since NIQE correlates poorly with subjective human opinion, we also adopt UNIQUE~\cite{9369977}. Smaller NIQE and larger UNIQUE indicate more naturalistic and perceptually favored quality. The NIQE and UNIQUE results on five datasets (DICM, LIME, MEF, VV, and NPE) are reported in Table~\ref{tab1}. 
For the NIQE and UNIQUE, SCL-LLE on two of the five datasets and has the best overall averaged results, indicating its across-scenes stability.
\subsubsection{Full-referenced image quality assessment.}
For full-reference image quality assessment, we employ the Peak Signal-to-Noise Ratio (PSNR,dB) and Structural Similarity (SSIM) metrics to compare the performance of different methods quantitatively. Since the five datasets used in the previous test contain no standard images, we use the Part2 of the SICE dataset~\cite{Cai2018deep} here. As shown in Table~\ref{Full}, our method obtained the best PSNR and SSIM among all the peer methods.
\begin{table}[!htb]
\centering
\caption{PSNR and SSIM on the Part2 of the SICE dataset.
}
\resizebox{1\columnwidth}{!}{
\begin{tabular}{c|cccccc|c}
\hline
Methods &LIME&R.Net&ISSR&Z.-DCE&E.GAN&RUAS&Ours\\ \hline
PSNR $\uparrow$& 13.67& 16.98  &15.01&14.78 &{17.82} &10.62 &\textbf{17.95}\\
SSIM $\uparrow$& 0.62& {0.66} &0.65 &0.62 &{0.66} &0.44 &\textbf{0.68}\\ 
\hline
\end{tabular}}
\label{Full}
\end{table}

\subsubsection{Human subjective survey.}
\begin{figure}[!]
\centering
\includegraphics[width=8.5cm]{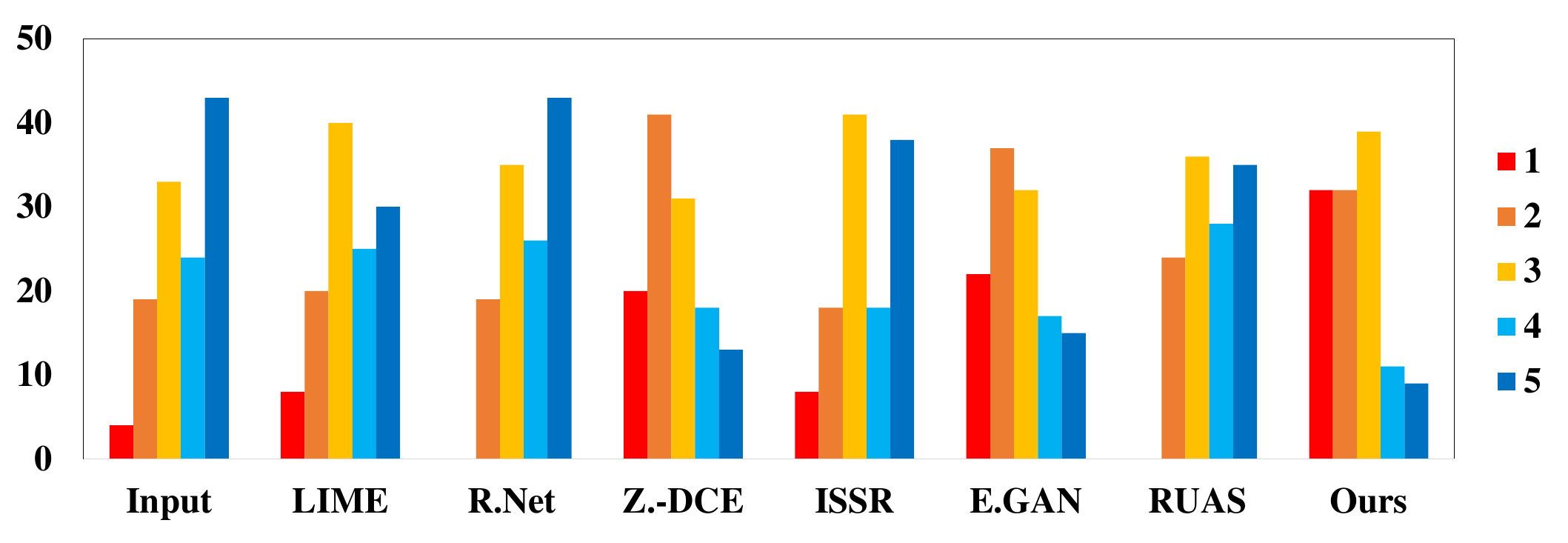}\\
\caption{The result in the human subjective survey. The color-changing from hot to cool means the quality transition from the best to worst; y-axis denotes the number of images in each ranking index.}
\label{US}
\end{figure}

We conduct a human subjective survey (user study) for comparisons. For each image in the five test datasets (DICM, LIME, MEF, VV, and NPE) enhanced by seven methods (LIME, Retinex-Net, Zero-DCE, ISSR, EnlightenGAN, RUAS, and our approach), we ask 11 human subjects to rank the enhanced images. These subjects are instructed to consider:\\
1) Whether or not the images contain visible noise.\\
2) Whether the images have over or underexposure artifacts.\\
3) Whether the images show non-realistic color or texture distortion.\\%
We stipulate the score of each image from 1 to 5, and the lower the value is, the better the image quality will be. 
The final results are shown in Table~\ref{tab1} and Fig.~\ref{US} 
, and our method is recognized as the best imaging quality.

\subsection{Ablation Study}
\subsubsection{Contribution of each loss.}
\begin{figure*}[h]
\centering
\subfigure[Input]{
\includegraphics[width=2.9cm]{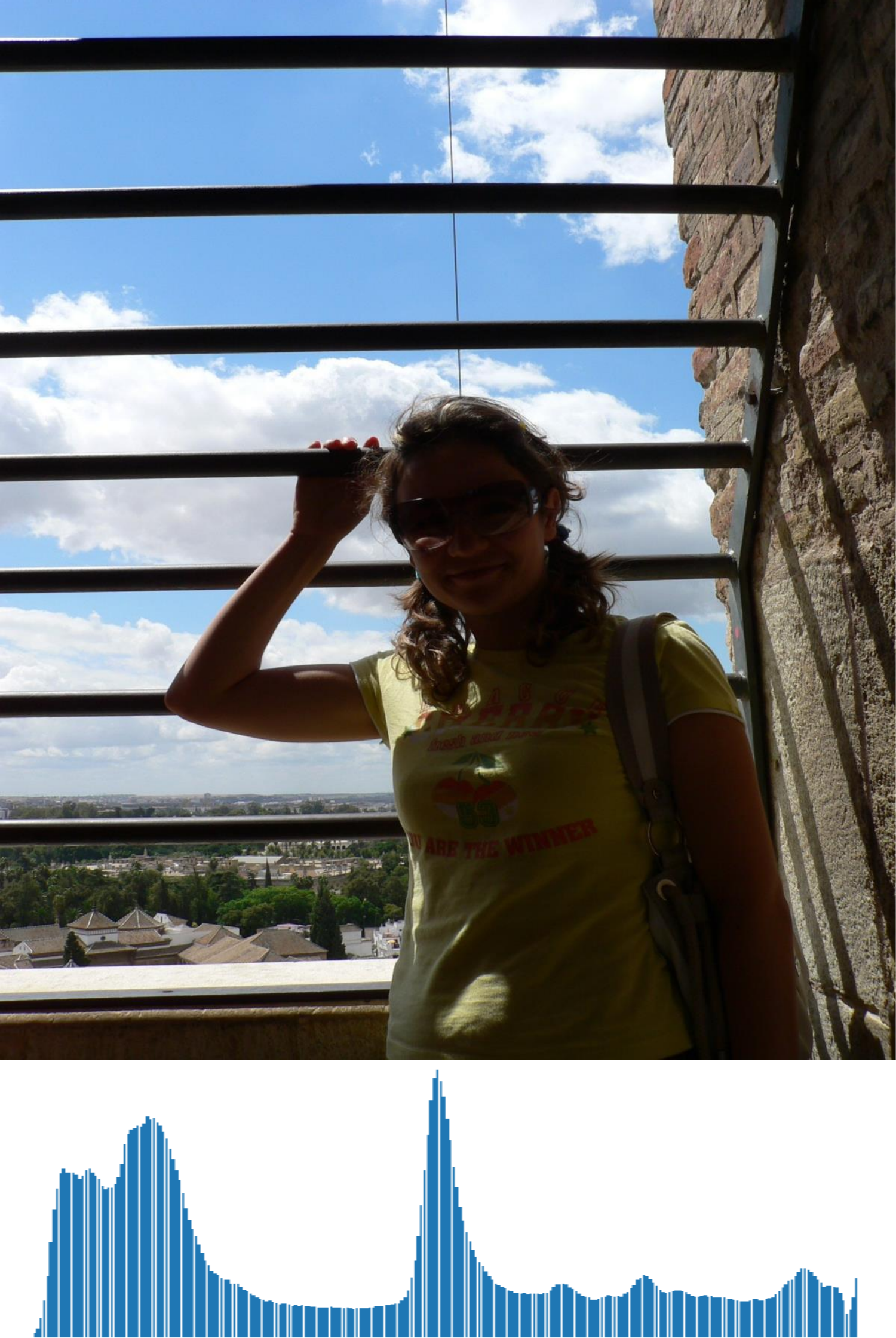}
}
\subfigure[w/o $L_{c}$]{
\includegraphics[width=2.9cm]{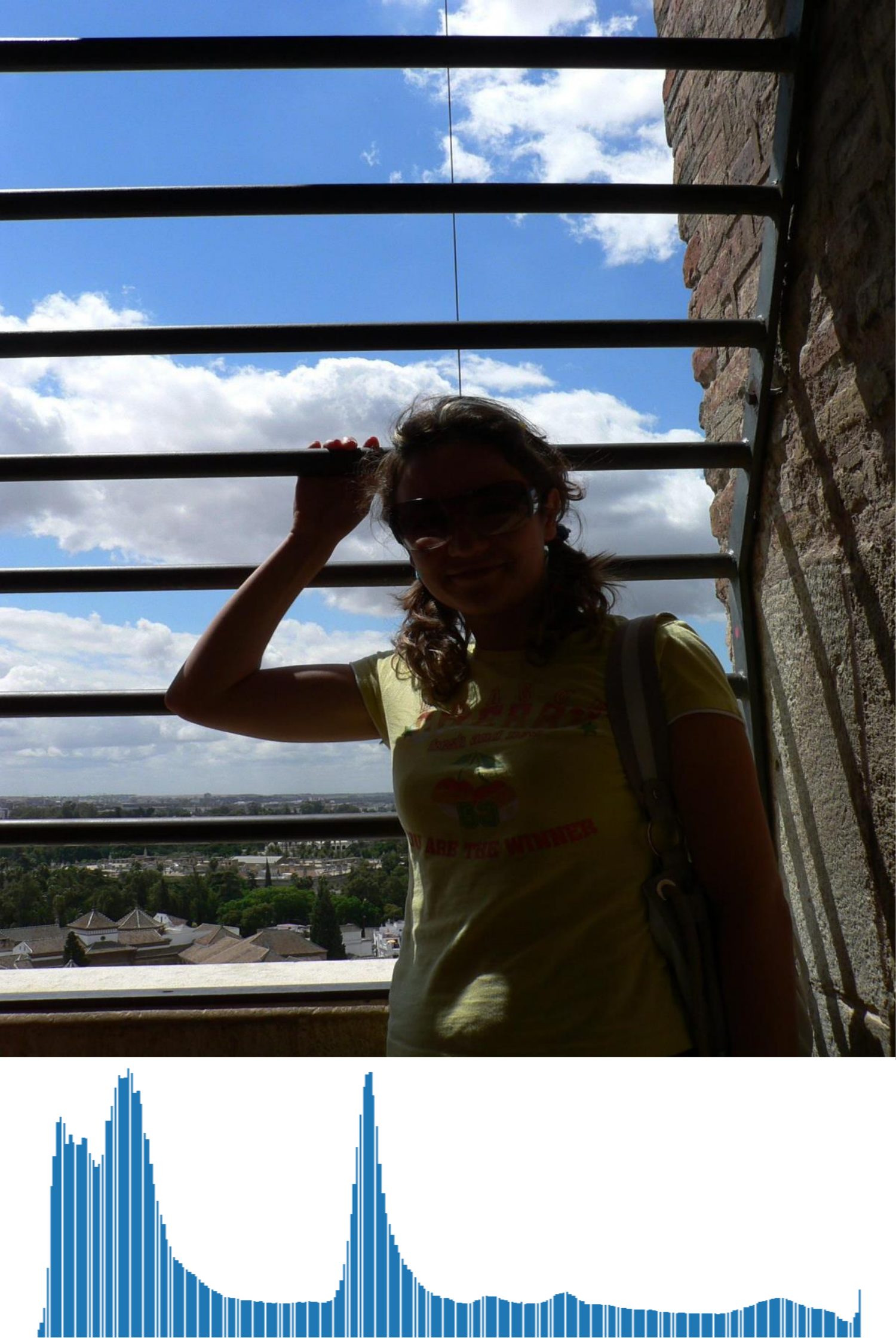}
}
\subfigure[w/o $L_{sc}$]{
\includegraphics[width=2.9cm]{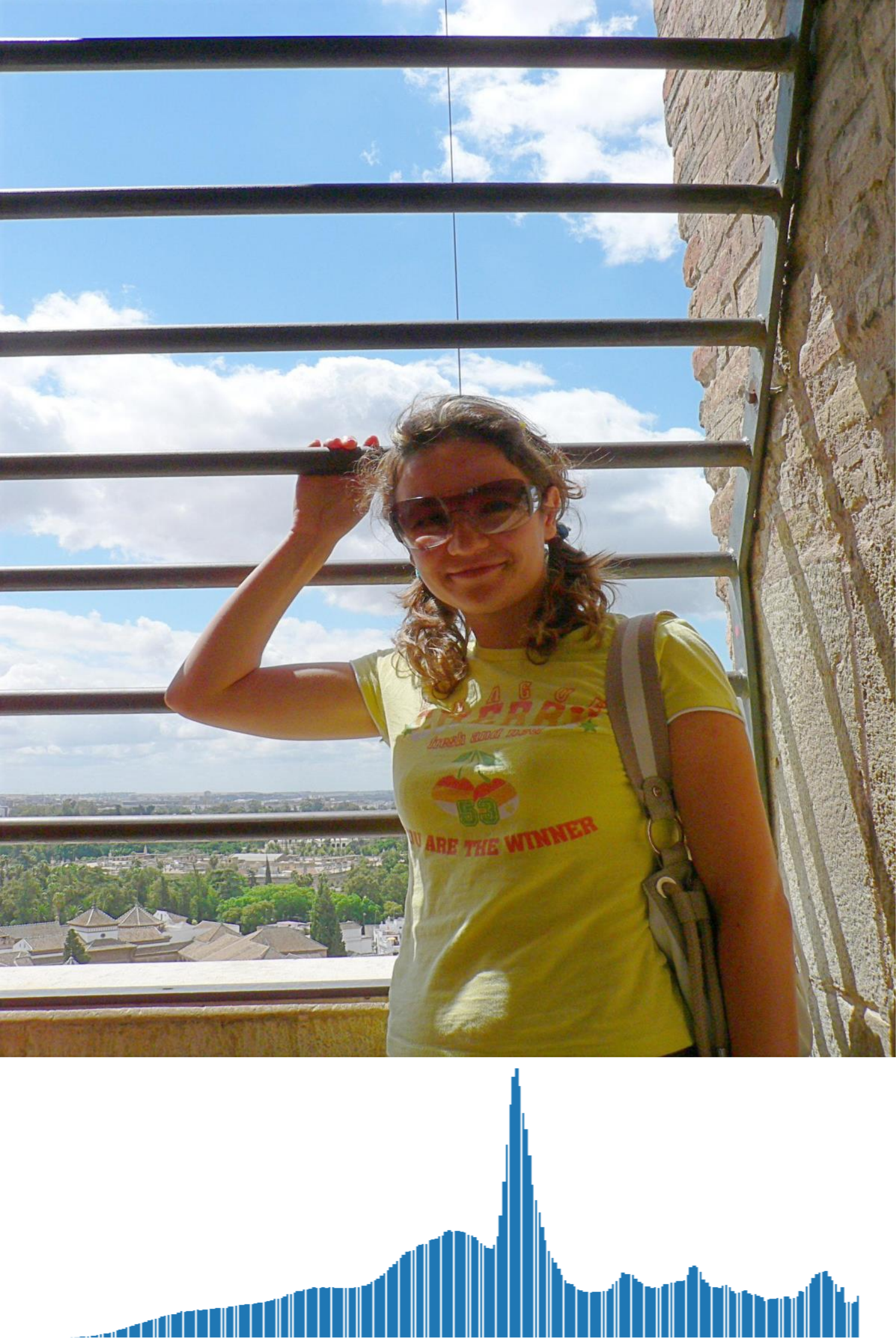}
}
\subfigure[w/o $L_{fr}$]{
\includegraphics[width=2.9cm]{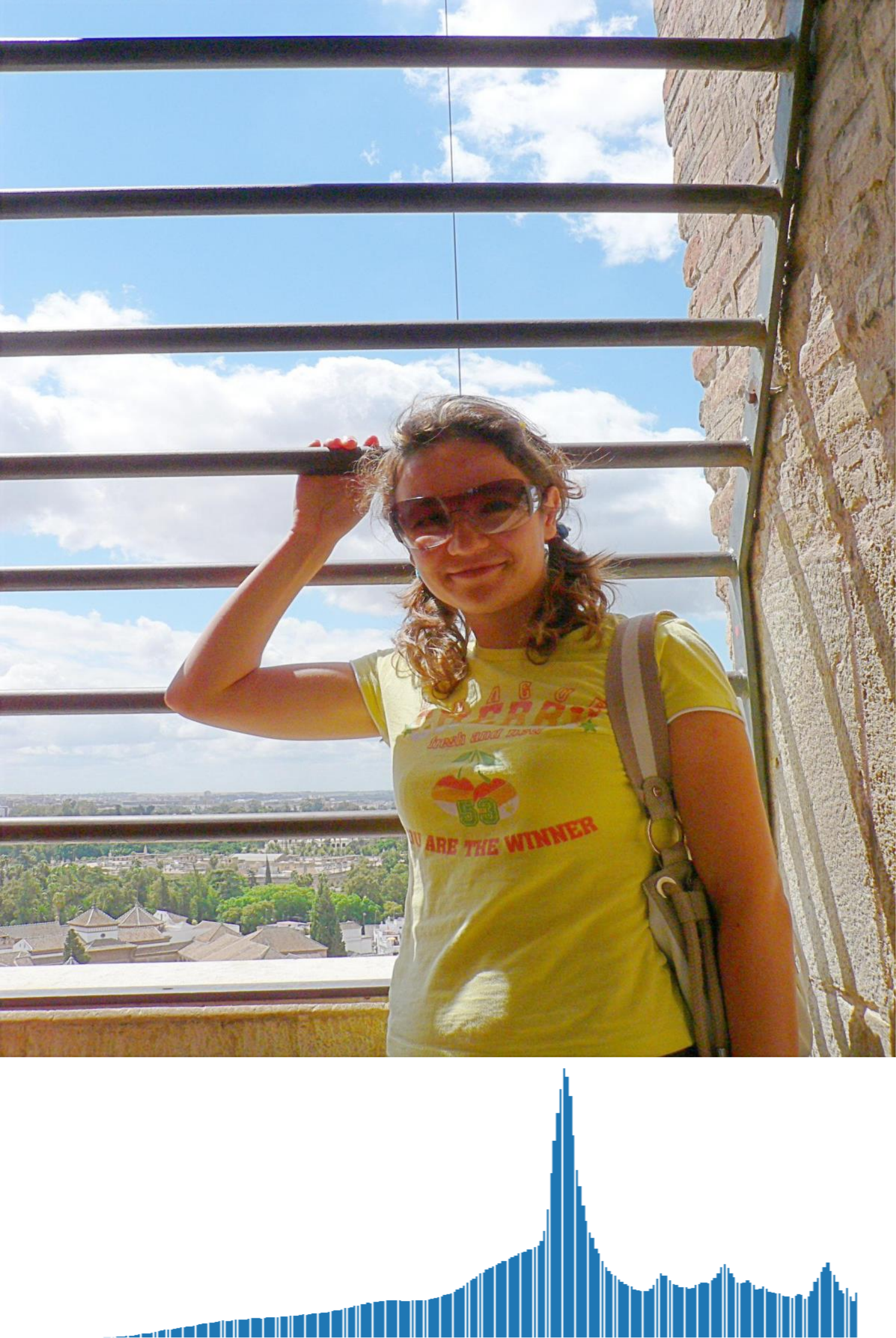}
}
\subfigure[Ours]{
\includegraphics[width=2.9cm]{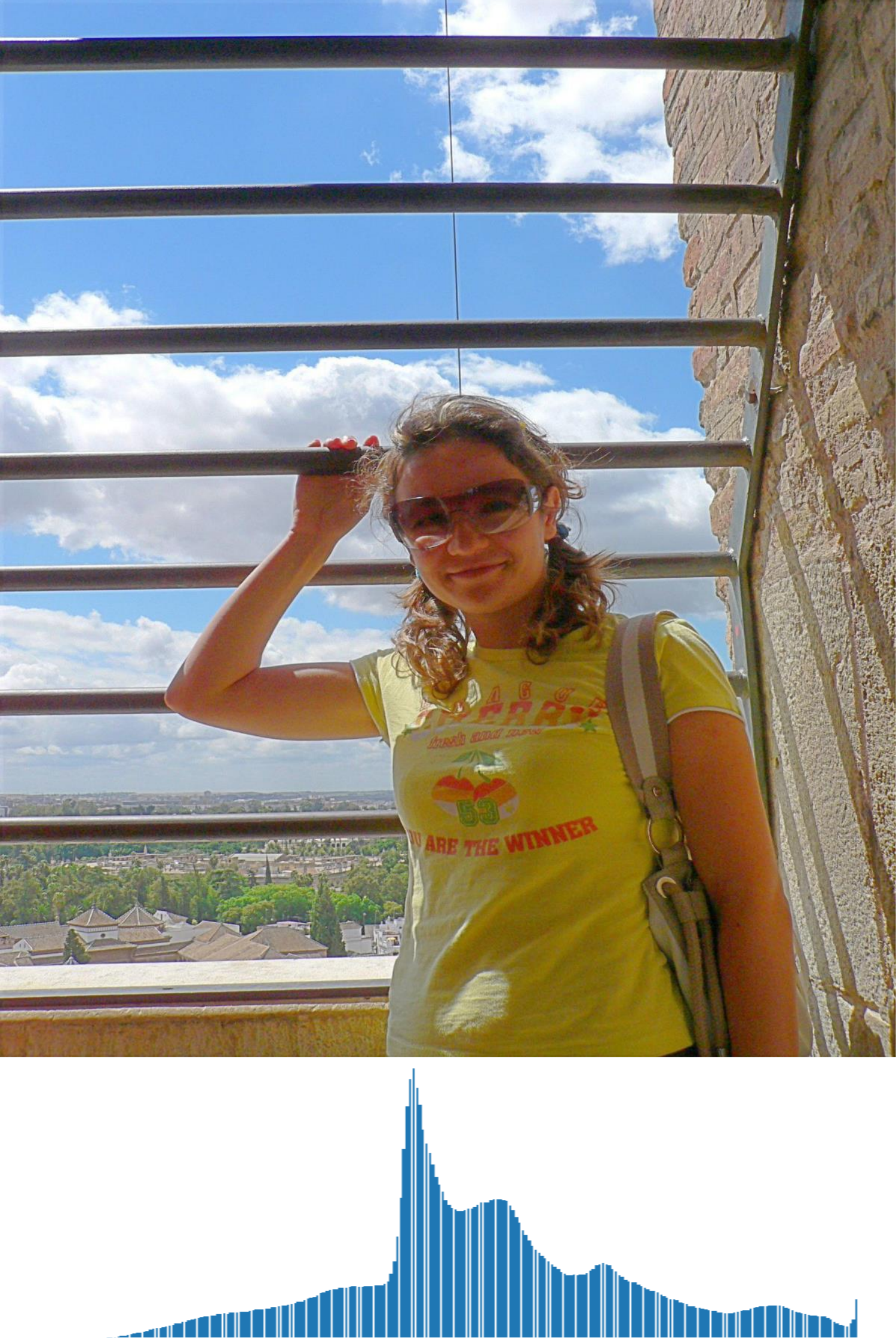}
}
\subfigure[Input]{
\includegraphics[width=2.9cm]{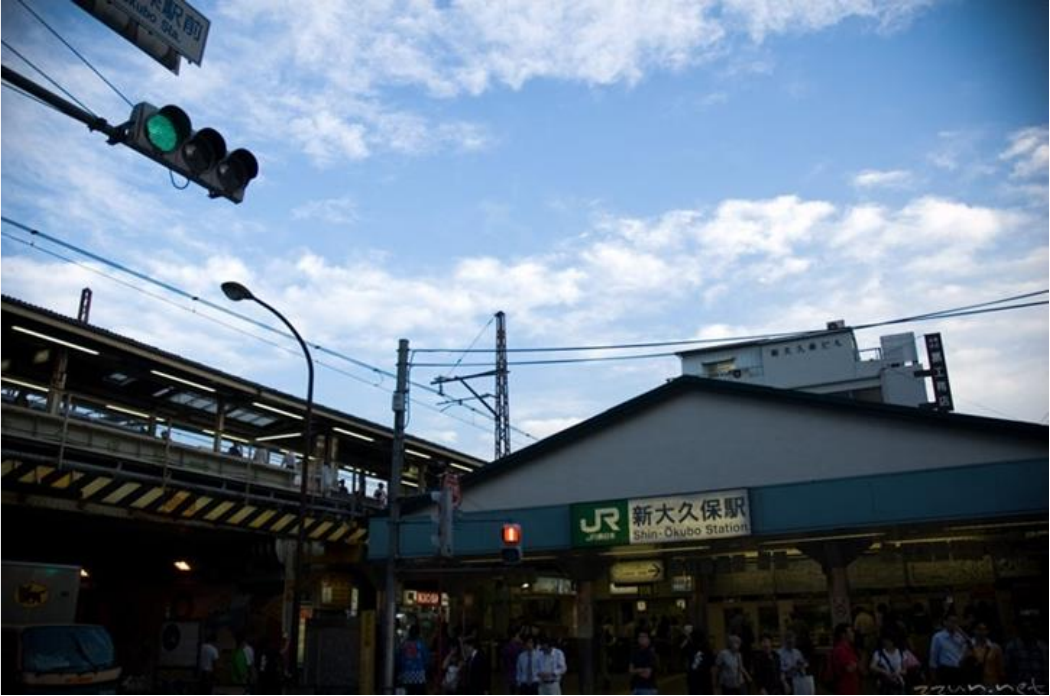}
}
\subfigure[w/o all neg. samples]{
\includegraphics[width=2.9cm]{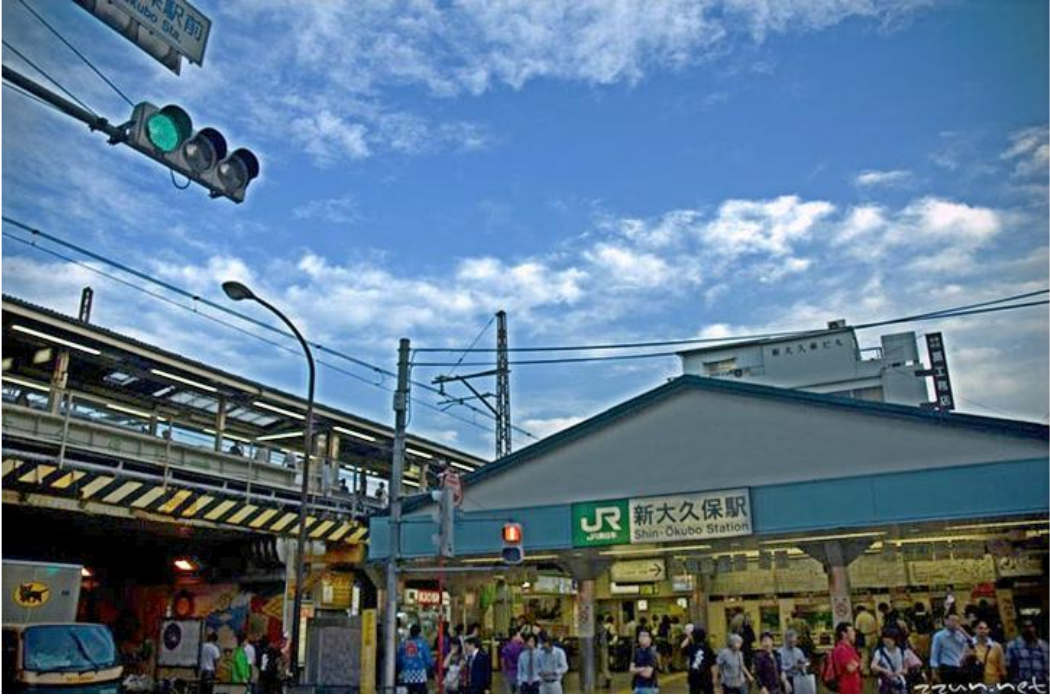}
}
\subfigure[w/o overexposed]{
\includegraphics[width=2.9cm]{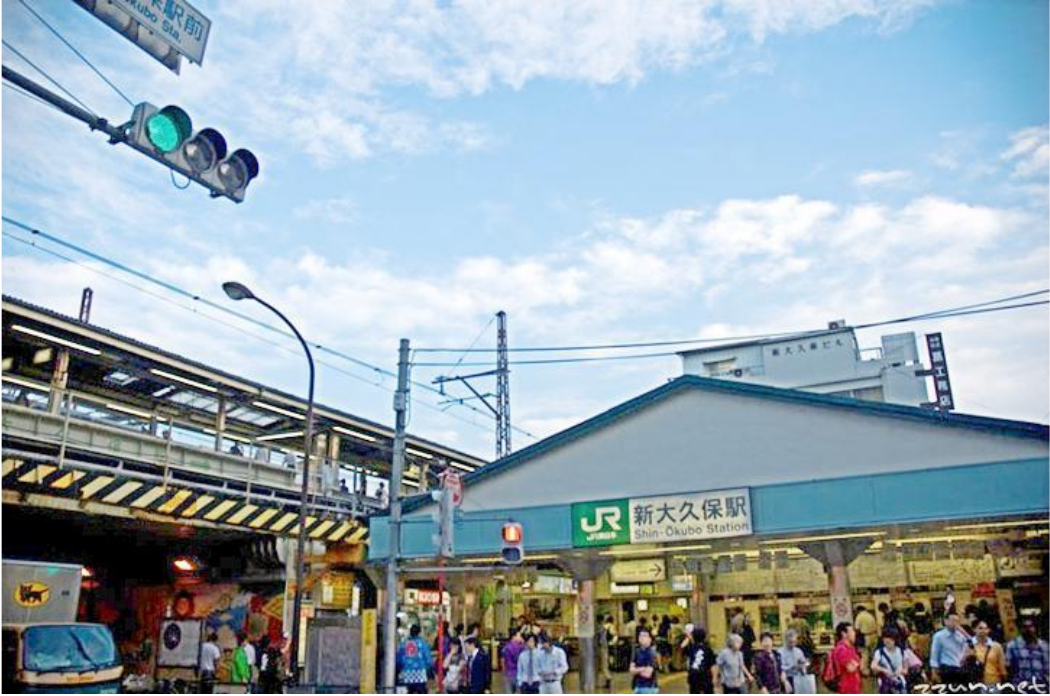}
}
\subfigure[w/o underexposed]{
\includegraphics[width=2.9cm]{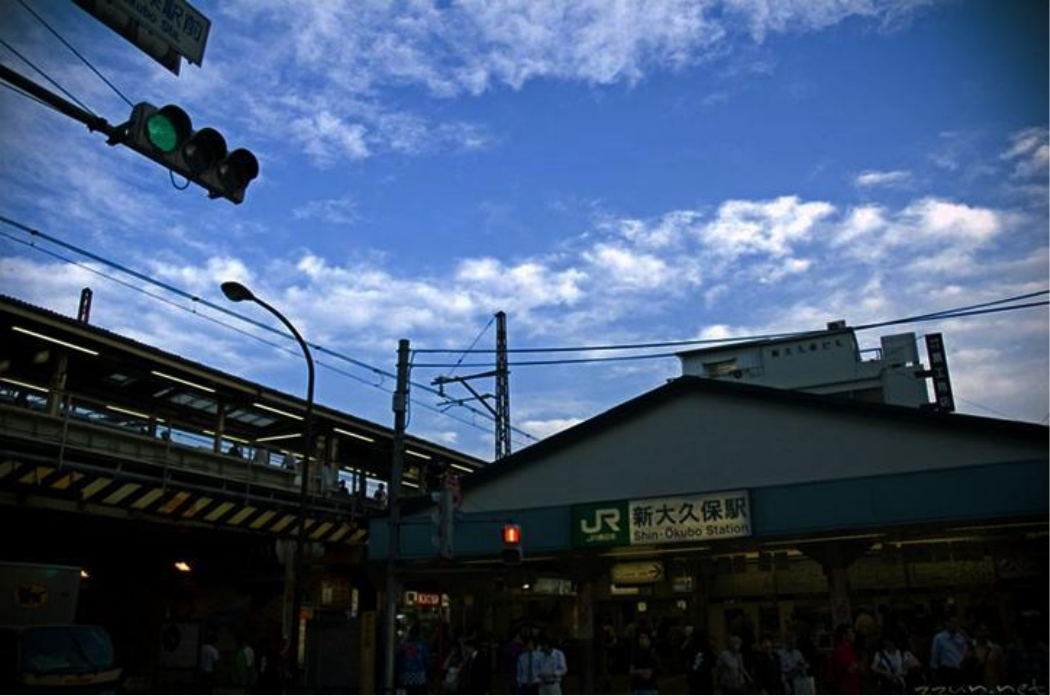}
}
\subfigure[Ours]{
\includegraphics[width=2.9cm]{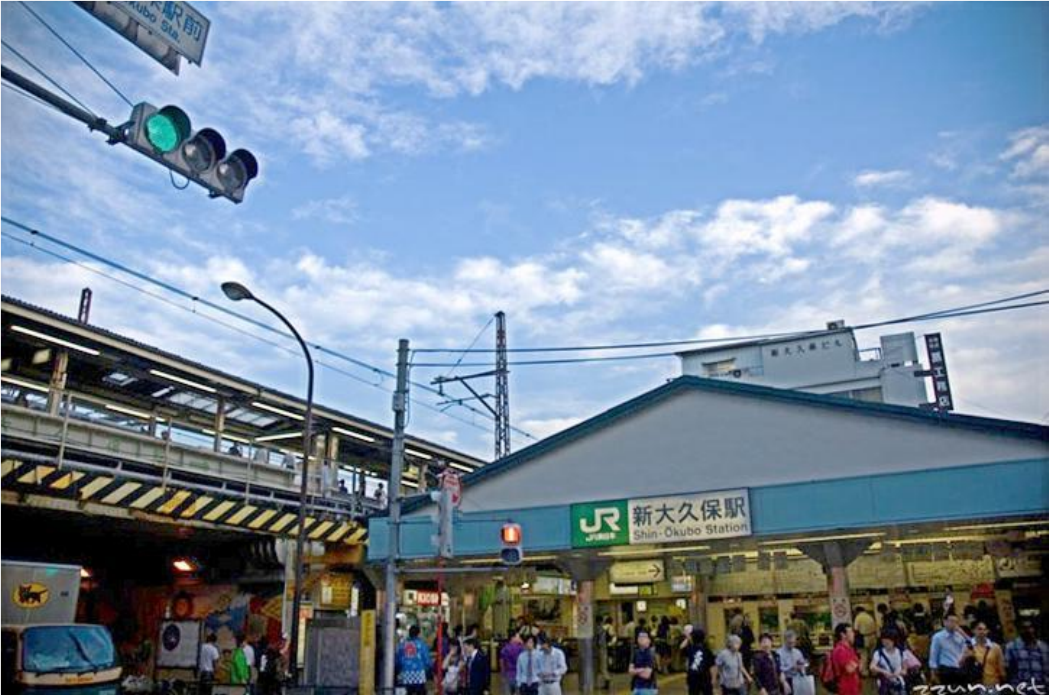}
}
\caption{Ablation study on the contribution of each loss and the impact of negative samples. 
}
\label{img2a}
\end{figure*}

We perform ablation studies to demonstrate the effectiveness of each loss component. Note that since the color consistency item $L_{cc}$ is initially proposed and tested in Zero-DCE~\cite{Guo_2020_CVPR}, we consider it as a baseline item without doing ablation study. Thus the feature preservation loss $L_{fp}$ is tested using the first item $L_{fr}$. 
The visualized samples with their corresponding histograms of the effects of $L_{c}$, $L_{sc}$ and $L_{fr}$ functions are shown in Fig.~\ref{img2a}~(a-e). Table~\ref{tab1a} shows the average NIQE and UNIQUE scores of each loss on five test sets. 
The contrastive learning loss $L_{c}$ plays a significant role in controlling the exposure level. The results without the semantic brightness consistency constraint loss $L_{sc}$ and without the feature retention loss $L_{fr}$ have relatively lower contrast ($e.g.$, the region of sky) than the final result. 
The overall losses enhance images with fine details and more naturalistic and perceptually favored quality. 
It can also be seen from the corresponding histograms that the final losses maintain a smooth mixture-of-Gaussian-like global distribution with rare over or under-saturation areas. Meanwhile, the undesirable unilateral over or under-saturation areas occur in the histograms of Fig.~\ref{img2a}~(b-d).

\begin{table}[!]
\centering
\caption{Ablation study. NIQE $\downarrow$ and UNIQUE (UN.) $\uparrow$ scores on the testing sets.}
\resizebox{1\columnwidth}{!}{
\begin{tabular}{c|cc|cc|cc|cc|cc|cc}
\hline
 &\multicolumn{2}{c|}{DICM} &\multicolumn{2}{c|}{LIME} &\multicolumn{2}{c|}{MEF} &\multicolumn{2}{c|}{VV} &\multicolumn{2}{c|}{NPE} &\multicolumn{2}{c}{Average} \\ 
\cline{2-13}
Methods &NIQE &UN. &NIQE &UN. &NIQE &UN. &NIQE &UN. &NIQE &UN. &NIQE &UN. \\ \hline
Input &4.26 &0.72 &4.36 &0.70 &4.26 &0.72 &3.52 &\textbf{0.74} &4.32 &\textbf{1.17} &4.13 &0.75\\ 
w/o $L_{c}$ &4.31 &0.64 &4.36 &0.57 &4.25 &0.56 &4.10 &0.70 &4.28 &1.02 &4.27 &0.66\\ 
w/o $L_{sc}$&{3.53} &0.83 &{3.85} &\textbf{0.76} &{3.32} &1.18 &3.21 &0.50 & 3.98 &1.02 &{3.49} &0.82\\
w/o $L_{fr}$ &3.54 &0.80 &3.88 &0.71 &{3.32} &1.22 &3.18 &0.47 &{3.97} &1.03 &3.50 &0.80\\
\hline        
w/o Neg. samples &3.55 &0.81 &{3.84} &0.72 &3.36 &1.14 &\textbf{3.14} &0.38 &{3.95} &1.01 &{3.49} &0.78\\
w/o overexp. Neg. &3.59 &0.75 &3.91 &0.59 &3.36 &1.24 &3.15 &0.37 &4.12 &0.87 &3.54 &0.74\\
w/o underexp. Neg. &4.58 &0.57 &4.52 &0.48 &4.69 &0.46 &3.58 &0.65 &4.36 &0.86 &4.38 &0.58\\
\hline                  
Ours  &\textbf{3.51} &\textbf{0.87} &\textbf{3.78} &\textbf{0.76} &\textbf{3.31} &\textbf{1.25} &{3.16} &0.49  &\textbf{3.88} &1.08 &\textbf{3.46} &\textbf{0.85}\\ 
\hline
\end{tabular}}
\label{tab1a}
\end{table}



\begin{table*}[!htp]
\centering
\caption{The segmentation results (mIoU $\uparrow$) of the input low-light image after enhancement on Cityscapes. 
}
\resizebox{1.7\columnwidth}{!}{
\begin{tabular}{c|cccc|cccc|cccc}
\hline
 &\multicolumn{4}{c|}{Frankfurt} & \multicolumn{4}{c|}{Lindau} &\multicolumn{4}{c}{Munster} \\
\cline{2-13}
Methods&$\gamma$=1& $\gamma$=2 & $\gamma$=6 &$\gamma$=10 & $\gamma$=1& $\gamma$=2 & $\gamma$=6 &$\gamma$=10 & $\gamma$=1& $\gamma$=2 & $\gamma$=6 &$\gamma$=10 \\ \hline
Input& \textbf{73.831}& \textbf{72.453}& 49.879& 33.577& \textbf{62.295}& \textbf{61.498}& 30.536& 17.894& \textbf{71.839}& 70.118& 45.519& 24.658\\
(TIP'17) LIME& 69.507& 70.361& 50.547& 36.928& 54.708& 59.259& 34.639& 16.877&{68.888} & 69.191& 53.132& 29.482\\
(BMVC’18) RetinexNet & 58.586& 56.471& 45.620& 29.463& 43.938& 44.943& 24.045& 13.725& 59.888& 56.118& 45.559& 22.785\\
(ACMMM’20) ISSR & 68.310& 70.297& 44.433& 30.866& 56.137& 59.919& 25.414& 16.973& 68.638& 69.907& 39.667& 23.176\\
(CVPR’20) Zero-DCE & 64.297& 69.127& {58.046}& {41.385}& 49.190& 56.599& {41.522}& {22.151}& 63.403& {70.153}& {55.300}& {33.109}\\
(TIP’21) EnlightenGAN &{70.669}& {70.624}& 50.757& 34.435& {57.242}&59.117&32.599&15.982& 68.671& 66.948& 48.965&29.182\\
(CVPR’21) RUAS & 41.374& 66.277& 51.890& 35.535& 41.482& 55.621& 36.501& 19.713& 53.152& 69.300& 52.015& 29.763\\ \hline
Ours & 67.437& 70.100& \textbf{58.488}& \textbf{42.047}& 51.909& {60.200}& \textbf{41.558}& \textbf{22.853}& 65.698& \textbf{70.710}& \textbf{56.379}& \textbf{33.442}\\ 
\hline
\end{tabular}}
\label{tab2}
\end{table*}

\subsubsection{Impact of negative samples.}
To verify the rationality of the negative sample selection in our experiment, we retrain SCL-LLE under different settings with:\\
1) 360 positive samples, \\
2) 360 positive and 360 underexposed negative samples, \\
3) 360 positive and 360 overexposed negative samples,  \\
4) 360 positive and 360 underexposed and 360 overexposed negative samples. \\
As shown in Fig.~\ref{img8}, Fig.~\ref{img2a}~(g) and Table~\ref{tab1a}, after removing the negative samples, SCL-LLE tends to use a biased white balance (color temperature) to enhance the well-lit regions, $e.g.$, the sky, the face, and the bird have turned darkened. Meanwhile, as shown in Fig.~\ref{img2a}~(h-i), removing the overexposed or underexposed negative samples leads to the white balance offset in the opposite direction. The above degraded results indicate the rationality and necessity of the usage of negative samples in our training framework.
\subsection{Semantic Segmentation with LLE}
Since there is no semantic annotation in current low-light image datasets, we use subsets Frankfurt, Lindau, and Munster in the validation set of Cityscapes to test the semantic segmentation performance before and after enhancement. In addition, we use the standard positive gamma transformation with a series of gamma values to simulate images with lower brightness. 
As shown in Table~\ref{tab2}, when using original input ($\gamma$=1), the semantic segmentation with all the enhancement models could not surpass the initial input. The reason may be in two aspects: 1) the current LLE methods, including ours, pay more attention to preserving visual-pleasing results, and more or less destroy the topological information on local areas; and 2) the normal-light scenes dominate the Cityscapes dataset. When $\gamma$ becomes larger, mIoU of segmentation after using the LLE methods has been significantly better than those using the original image. Among all the methods, the segmentation performance with our method tends to be the best when the scene tends to be dark. These results encourage us to bridge the gap between the current LLE methods and the downstream tasks.

\section{Conclusion}
We have proposed an effective semantically contrastive learning paradigm (SCL-LLE) to solve the low-light image enhancement problem. SCL-LLE reveals how unpaired negative and positive samples can produce visual-pleasing images, and illustrates how we can leverage semantic information to preserve a visual similarity between the input and the output. 
We cast image enhancement as multi-task joint learning, where SCL-LLE is converted into three constraints of contrastive learning, semantic brightness consistency, and feature preservation for simultaneously ensuring the color, texture, and exposure consistency. Experiments exhibit clear improvements of our method over existing state-of-the-arts LLE models on six cross-domain datasets. 
Meanwhile, experiments also reveal that our SCL-LLE has the potential to guide the downstream semantic segmentation task to gain better performance.


\section{Acknowledgments}
We would like to thank Prof. Songcan Chen and Sheng-Jun Huang from NUAA, and Dr. Dong Zhang from NJUST for their important suggestions. This work was supported by AI+ Project of NUAA (XZA20003), Natural Science Foundation of China (62172218, 61772268, 62172212), Natural Science Foundation of Jiangsu Province (BK20190065).

\bibliography{Semantically_Contrastive_Learning_for_Low-light_Image_Enhancement}

\end{document}